	\newif\ifboxes
	\documentclass[a4paper, 10 pt, conference]{ieeeconf}  % Comment this line out if you need a4paper

	\IEEEoverridecommandlockouts                              % This command is only needed if 
	\title{\LARGE \bf
	Sobi: An Interactive Social Service Robot for Long-Term Autonomy\\ in Open Environments
	}

	\author{Marvin Stuede, Konrad Westermann, Moritz Schappler and Svenja Spindeldreier$^{1}$% <-this % stops a space
	\thanks{$^{1}$All authors are with the Leibniz University Hannover, Institute of Mechatronic Systems, D-30823 Garbsen, Germany,
	        {\tt\small marvin.stuede@imes.uni-hannover.de}
        	\newline 978-1-6654-1213-1/21/\$31.00 \textcopyright 2021 IEEE}%
	}

	\newcommand{\capsize}{\fontsize{\small}\selectfont}
	\usepackage[compact]{titlesec}
	\usepackage[nolist]{acronym}
	\usepackage{cite}
	\usepackage{pdfpages}
	\usepackage{adjustbox}
	\usepackage{framed}
	\usepackage[ruled]{algorithm2e}
	\usepackage{nicefrac}
	\usepackage{amsmath}
	\usepackage[hidelinks]{hyperref}

	\usepackage{amssymb}
	\usepackage{subcaption}
	\usepackage[font=small]{caption}
	\usepackage[T1]{fontenc}
	\usepackage[utf8]{inputenc}
	\usepackage{grffile}
	\usepackage{siunitx}
	\usepackage{adjustbox}
	\usepackage{setspace}
	\usepackage{dblfloatfix} 
	\let\labelindent\relax
	\usepackage{enumitem}
	\usepackage{makecell}
	\usepackage{verbatim}
	\usepackage{forest}
	\usepackage{tikz}
	\usepackage[english]{babel}
	\usepackage{blindtext}
	\usepackage{multirow, bigdelim, booktabs}
	 \usepackage{blkarray} 
	 \usepackage{graphicx}
	\usepackage{array}
	\usetikzlibrary{tikzmark}
	\usetikzlibrary{shadows,arrows.meta, shapes, fit, calc, positioning, backgrounds}
		\definecolor{bt-blue}{RGB}{2,170,255}
	\tikzset{parent/.style={align=center,text width=2cm},
		child/.style={align=center,text width=2cm},
		grandchild/.style={text width=2.3cm},
		edges/.style={for tree={parent anchor=south, child anchor=north,align=center,edge={->},base=bottom,where n children=0{tier=word}{}}}, 
		bt-action/.style={top color=green!10, bottom color=green!20,draw},
		bt-subtree/.style={top color=bt-blue!10, bottom color=bt-blue!20,draw},
		bt-condition/.style={ellipse,top color=orange!10, bottom color=orange!20,draw,minimum width=2.0cm, minimum height=2mm},
		bt-root/.style={minimum size=0.3cm, forbidden sign,draw , fill=white},
		bt-sequence/.style={shape=rectangle, minimum size=0.6cm, font={\textrightarrow},draw , fill=white},
		bt-fallback/.style={shape=rectangle, minimum size=0.6cm, font={?},draw , fill=white},
		bt-etc/.style={minimum width=1.8cm, font={\dots}, general shadow/.style=},
	}

	\pgfdeclarelayer{bg}    % declare background layer
	\pgfdeclarelayer{c1}
	\pgfdeclarelayer{c2}
	\pgfdeclarelayer{c3}
	\pgfsetlayers{bg, c1, c2, c3, main}

	\usepackage{pgfplots} % pgfplots zum plotten in LaTeX
	\pgfplotsset{compat=1.3,
		tick label style={font=\small,/pgf/number format/use comma}, %Komma nutzen in allen Axen
		every axis legend/.append style={at={(0.5,1.03)},anchor=south,nodes=right,font=\small}, % Legende 
		label style={font=\capsize} 
	}
	\newlength\picwidth
	\newlength\boxwidth
	\newlength\tikzwidth
	\newlength\figureheight
		\newlength\sfwidth
	\newlength\figurewidth
	\newlength\shorten
			\DeclareSIUnit{\amperehour}{Ah}
			\newif\ifcopyright
	\copyrighttrue
	
\begin{document}
			\ifcopyright
		{\LARGE IEEE Copyright Notice}
		\newline
		\fboxrule=0.4pt \fboxsep=3pt
		
		\fbox{\begin{minipage}{1.1\linewidth}  % <-- hier Kastenbreiter der Kopfzeile ändern
				%\centering
				% <-- hier Namen der Konferenz und Jahr einfügen
				%Changes were made to this version by the publisher prior to publication. \\
				%	The final version of record is available at http://doi.org/10.1109/XYZ123456789.TODO  % <-- hier DOI einfügen
				Copyright (c) 2021 IEEE. Personal use of this material is permitted. For any other purposes, permission must be obtained from the IEEE by emailing pubs-permissions@ieee.org. \\
				
				Accepted to be published in: Proceedings of the 10th European Conference on Mobile Robots (ECMR 2021), August 31st – September 3rd, Bonn, Germany
				
		\end{minipage}}
		\else
		\fi
	\begin{acronym}
		\acro{bt}[BT]{Behavior Tree}
		\acro{hri}[HRI]{human-robot interaction}
		\acro{ros}[ROS]{Robot Operating System}
		\acro{tsl}[TSL]{total system lifetime}
		\acro{ntp}[NTP]{Network Time Protocol}
		\acro{lta}[LTA]{long term autonomy}
		\acro{gqs}[GQS]{Godspeed Questionnaire Series}
	\end{acronym}
	\newcommand{\eal}{et al.}

	\newcommand{\ealcite}[1]{\eal~\cite{#1}}
	\newcommand{\TODO}[1]{\textcolor{red}{\textbf{TODO: #1}}}
	\renewcommand{\vec}[1]{\mbox{\boldmath{$#1$}}}
	\newcommand{\dvec}[1]{\dot{\mbox{\boldmath{$#1$}}}}
	\newcommand{\ddvec}[1]{\ddot{\mbox{\boldmath{$#1$}}}}
	
	\newcommand{\transpose}{\ind{T}}
	
	\newcommand{\tmat}[2]{{^{\ind{#1}}\vec{T}_{\ind{#2}}}}
	\newcommand{\rmat}[2]{{^{\ind{#1}}\vec{R}_{\ind{#2}}}}
	\newcommand{\ks}[1]{{\ind{(KS)_{#1}}}}
	\maketitle
	\thispagestyle{empty}
	\pagestyle{empty}

	%%%%%%%%%%%%%%%%%%%%%%%%%%%%%%%%%%%%%%%%%%%%%%%%%%%%%%%%%%%%%%%%%%%%%%%%%%%%%%%%
	\begin{abstract}
Long-term autonomy in service robotics is a current research topic, especially for dynamic, large-scale environments that change over time. 
We present Sobi, a mobile service robot developed as an interactive guide for open environments, such as public places with indoor and outdoor areas. The robot will serve as a platform for environmental modeling and human-robot interaction.
Its main hardware and software components, which we freely license as a documented open source project, are presented.
Another key focus is Sobi's monitoring system for long-term autonomy, which restores system components in a targeted manner in order to extend the total system lifetime without unplanned intervention.
We demonstrate first results of the long-term autonomous capabilities in a 16-day indoor deployment, in which the robot patrols a total of 66.6 km with an average of 5.5 hours of travel time per weekday, charging autonomously in between.
In a user study with 12 participants, we evaluate the appearance and usability of the user interface, which allows users to interactively query information about the environment and directions.
	\end{abstract}

	%%%%%%%%%%%%%%%%%%%%%%%%%%%%%%%%%%%%%%%%%%%%%%%%%%%%%%%%%%%%%%%%%%%%%%%%%%%%%%%%
	\section{Introduction}

	Social service robots can now be found performing tasks such as providing information, guiding, transport or entertainment in various contexts, e.g. retail, airports \cite{Triebel2016} or hotels \cite{Ivanov2017}.
	Despite their commercial availability,  \ac{lta} without human intervention, especially in unstructured and dynamic environments, remains a current research challenge.
	Long-term autonomous operations thereby pose special demands to robot design, both in terms of robust and reliant software and hardware.
	
\begin{figure}[t]
		\includegraphics[width=1\linewidth]{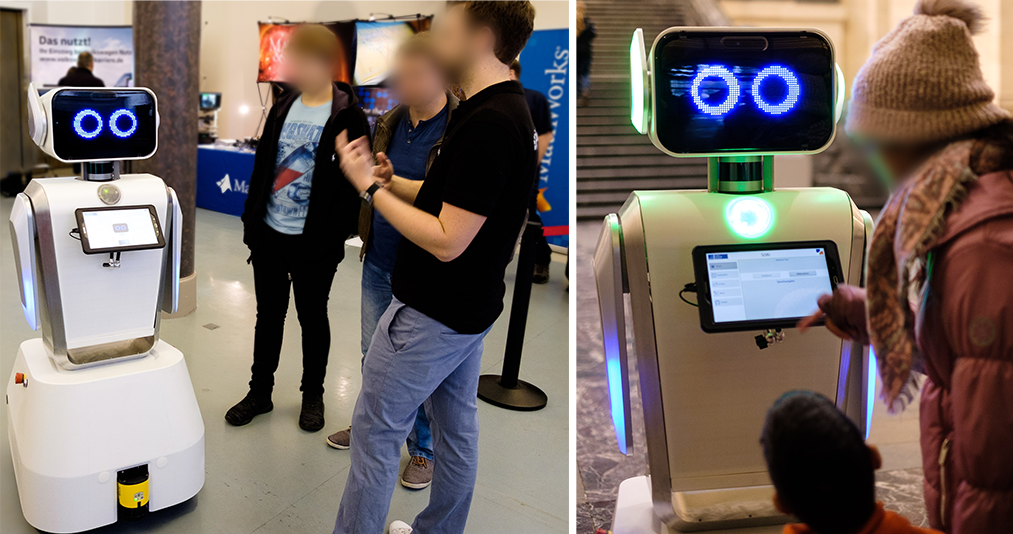}
		\caption{Sobi interacting with users during one of the presentations at public events (Hannover city hall).}
		\label{fig:real_photo}
		\vspace{-7mm}
	\end{figure}

	In this paper we present Sobi, a social service robot for information provision and guiding in open environments (see Fig.~\ref{fig:real_photo}).
	Sobi is designed for use on a campus inside and outside buildings and answers voice- and touch-based requests for e.g. directions, room plans, canteen menus or small talk  and provides a guiding functionality.
	
	The robotic tour guide and information terminal scenario is a popular use case and has been considered in various contexts in research.
	Early developments focused mostly on robust localization and navigation, such as the RHINO robot which was deployed for 6 days in a museum \cite{Burgard1998}.
	Later implementations reached cumulative operating times of several weeks. 
	Sacarino is a service robot that was deployed for multiple weeks to provide information and do navigation tasks for guests in a hotel \cite{Pinillos2016}.
	In the CoBot project, cumulative travelled distances of more than \SI{1000}{\kilo\meter} were achieved with multiple robots \cite{Biswas2016}.
	The project focused on long-term mapping, navigation and \ac{hri}, especially to proactively ask humans for help in problematic situations (e.g. operating a lift without physical manipulators).
	Between autonomous operations, however, manual intervention by supervisors is necessary, for example for the charging process.
	In \cite{Meeussen2011}, it is described that autonomous charging capabilities as well as deliberate requests for supervisor assistance can significantly lengthen the deployment time.
	The EU project STRANDS focused on the development of specific methods for environmental modeling and \ac{hri} for \ac{lta} operations.
	Their robots reached multiple weeks of uninterrupted autonomy in security and care contexts \cite{Hawes2016} and tour-guide scenarios \cite{Duchetto2019} in subsequent works.
	Other projects focus on types of interaction e.g. by voice and facial recognition \cite{DeJong2018}\cite{Wang2018}, specific environments (e.g. outdoors \cite{Chiang2008} or elder care \cite{Luperto2019}), or perception and social navigation \cite{Triebel2016}.
	
	However, research is often based on the use of commercially available robots, usually supplemented by only a few hardware components, or the systems are designed for either exclusive indoor  \cite{Biswas2016, Hawes2016, Duchetto2019, Pinillos2016, Meeussen2011, Wang2018, Luperto2019} or outdoor use \cite{Chiang2008}, and the used sensors and actuators would make them unsuitable for the respective other area.
	In contrast, Sobi is aimed at long term autonomous use in unsupervised open spaces, multiple days and weeks indoors and frequent partially supervised deployments outdoors.
		%One of the  main focuses of Sobi is long-term autonomous (\acs{lta}\acused{lta}) use in unsupervised open spaces, multiple days and weeks indoors and frequent partially supervised deployments outdoors.
	Contrary to other \ac{lta} systems, we provide not only the complete robot's software as open source, but also make the system available as an open source hardware project, including 3D-model and design data, mechanical and electrical drawings and parts lists\footnote{\url{https://marvinstuede.github.io/Sobi}}.
	%In addition, the complete software of the robot as well as the hardware  are freely useable and available as open source 
	
	One key aspect of \ac{lta} is monitoring system variables and executing defined recoveries in the event of a failure, which has been explored particularly for areas outside of service robotics, such as underwater or extraterrestrial applications.
	This can involve model-based methods that compare the system and component behavior with a nominal target \cite{Antonelli2014} or data-driven approaches for outlier rejection \cite{Khalastchi2018}.
	Although various descriptions for \ac{lta} applications in service robotics note that monitoring is an important aspect \cite{Hawes2016, Meeussen2011, Wang2018}, the specific methods used are not presented.
	
	Therefore, this paper firstly introduces the main design and components of Sobi, with an emphasis on \ac{lta} aspects such as sensor and algorithmic setup for robust localization indoors and outdoors.
	Furthermore, a novel type of stateless reactive monitoring system based on \acp{bt} as part of Sobi is presented, which detects faults and reacts accordingly by restarts or recoveries.
	In the course of the evaluation, we show first results on \ac{lta} with a 16-day deployment in which the robot patrols a total of \SI{66.6}{\kilo\meter} within a building and notifies people of compliance to hygiene regulations.
	The external appearance as well as the human-robot interface are examined in a user study, that includes verbal and touch requests for places of interest, public transport and meal plans.
	To summarize, the contributions of this paper are
	1) a novel social service robot under free license for the use in indoor and outdoor environments,
	2) a stateless reactive monitoring system based on \acp{bt} for \ac{lta} of the robot,
	3) experimental evaluation of the robot in a scenario aiming at \ac{lta} and in a user study.
	
	The remainder of this paper is structured as follows: The next section \ref{sec:robot} gives on overview of the main hardware and software components of Sobi.
	The monitoring system is described in section \ref{sec:monitoring}.
	Subsequently, section \ref{sec:evaluation} describes the long-term test and user study.
	In section \ref{sec:lessons} we give some remarks about the lessons learned throughout the development process.
	Finally, in section \ref{sec:conclusion} we provide a summary and an outlook on further work.

    \section{Robot design and components}
    	\label{sec:robot}
    This section describes the design, hardware and software components of Sobi, starting with a brief overview of the core requirements:
    
    The robot is supposed to operate in various buildings as well as in the outdoor area of the newly built campus of the Faculty of Mechanical Engineering of the Leibniz University Hannover.
	Within the robot's home building, fully autonomous operation should be possible by specifically using a charging station when needed.
    In other areas, the robot should operate autonomously as long as the batteries are charged and will then be manually transported back to the home building.
    It should be able to operate for several hours with robust localization and navigation in these dynamic environments, and also enable multi modal perceptual sensing for applications such as person recognition and tracking.
    Since the robot should look appealing, one of the development priorities is an approachable outer design, which at the same time protects the robot's components from environmental influences,  and an intuitive user interface.
    The robot should therefore have basic splash water protection to withstand spills or brief drizzles (according to the IP21 protection class to protect against touch and vertically falling water).
	However, it should be noted that we used the IP rating only as a guideline and no industrial grade tests were performed.
   
  \subsection{Exterior Shape and Design Concept}
  The shape and color concept of the robot was developed in close cooperation with the Hanover University of Applied Sciences and Arts.
  Sobi's design is based on a futuristic appearance and clear lines, which is intended to have a friendly and inviting effect on users.
  The shape is based on simple geometric bodies offset with chamfers, which can be found throughout the design of the robot.
  The bevelled design and the use of straight lines are furthermore intended to avoid water build-up and to ease sealing of transition points.
  One individual design feature is the circumferential aluminum-bracket that connects the upper body to the base and was CNC-machined.
  Since the robot is relatively large with a  height of \SI{1.56}{\meter} and width of  \SI{0.66}{\meter}, this feature is supposed to give an impression of lightness.  
  Humanoid features are the torso and head with eyes, which are indicated as rings, as well as movable arms and ears, which can be illuminated in color by LEDs.
  %The arms and ears also have flat, circumferential chamfers and appear to be free-floating due to their distance from the body.
  Flat elements (e.g. for arms and ears) were laser-cut from aluminum and ABS plastic.
   The outer white covers of the robot, with the exception of the base cover, are laser sintered and  coated to repel water.
%Therefore, the electrical components are enclosed to achieve basic splash protection in the event of a spill or brief drizzle.
\subsection{Hardware Components}
The described externally visible elements of the robot are supported by a structure made of aluminum square-profiles, which is rigidly connected to a \textbf{wheeled platform}.
All components mentioned in this section are shown in  Fig.~\ref{fig:robot}.
The platform (Neobotix MP-500: differential drive with one caster wheel) has an onboard computer and \textbf{2D-Lidar}, which can trigger low-level emergency stops for collision avoidance.
It includes two \SI{12}{\volt} AGM batteries in series with a total capacity of \SI{50}{\amperehour}, which we extended by two smaller batteries for a total capacity of \SI{75}{\amperehour}.
This gives a minimum time of 4--5 hours until a recharge is needed for heavy usage and over twelve hours standby time.

The embedded \textbf{main computer} used for controlling the robot is mounted on the platform (Vecow EVS-1010: Intel i7-7700T, 16 GB RAM, GeForce GTX1050 GPU).
It has two built-in WiFi modules to be permanently connected to the internet and to provide a WiFi hotspot for external access in field operation.
The robot includes the following sensors for localization, navigation and interaction:
			\begin{figure*}[t]
	
	\begin{minipage}[t]{0.6\textwidth}
		
		\includegraphics[width=1\textwidth]{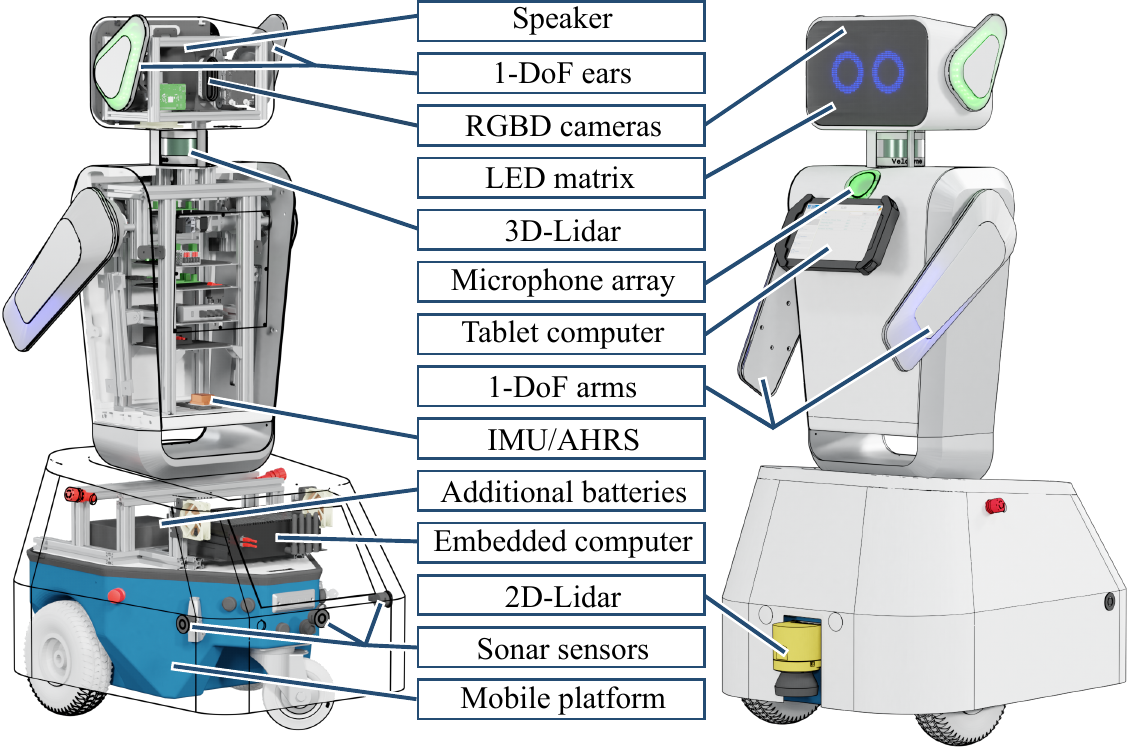}
		\caption{Hardware components and their placement in Sobi.}
		\label{fig:robot}

	\end{minipage}
	%\hspace{1cm}
	\begin{minipage}[t]{0.39\textwidth}
		\newlength{\hmargin}
		\setlength{\hmargin}{6mm}
	%	\hspace{\hmargin}
	\centering
		\includegraphics[width=0.81\textwidth]{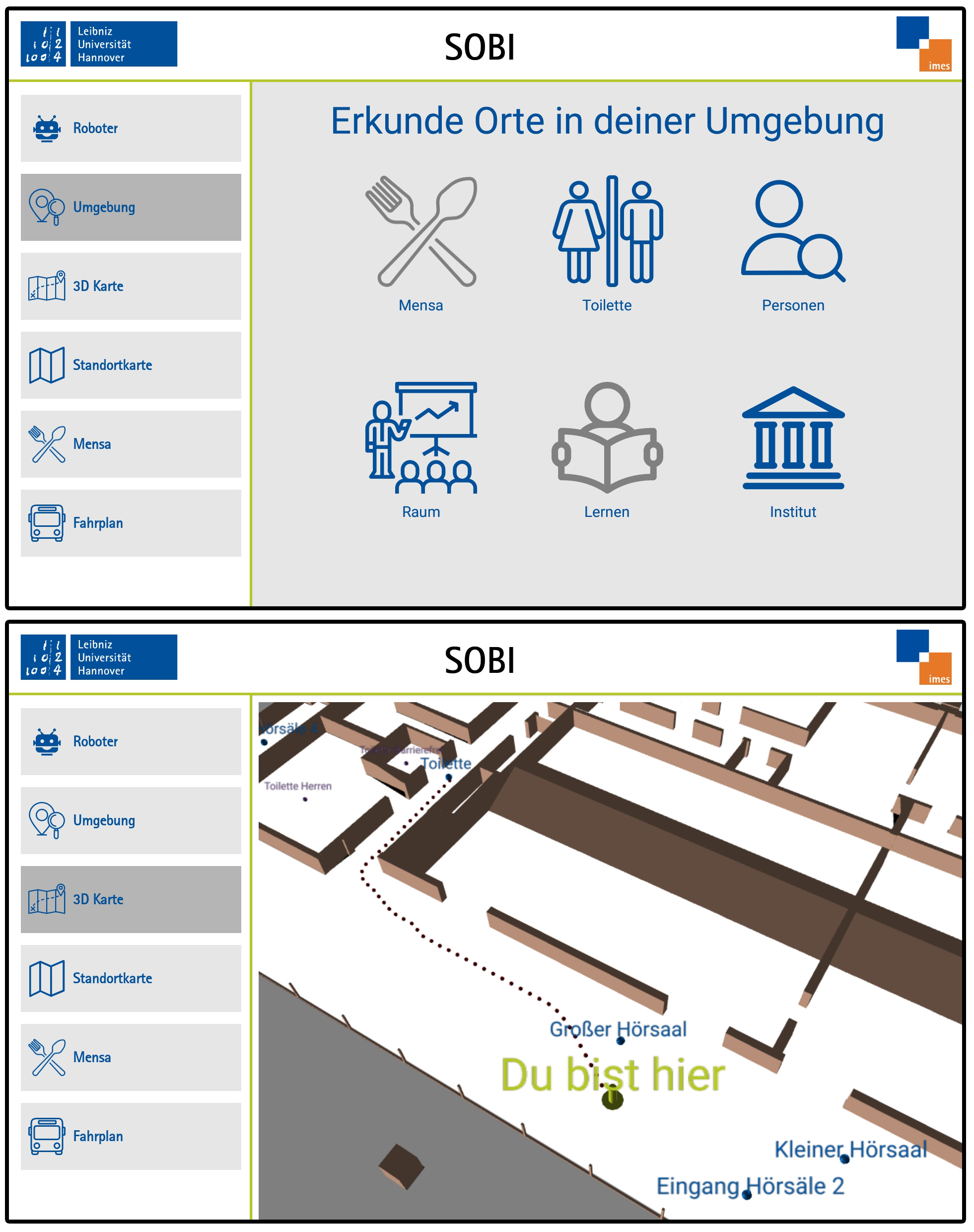}
		%	\hspace{\hmargin}
		\caption{GUI of the tablet. Places of interest (POI) can be chosen by category (top). A 3D map then shows the path from the current position (bottom).}
		\label{fig:gui}
	\end{minipage}
	\vspace{-5mm}
\end{figure*}
\begin{itemize}
	\item Xsens MTi-30 \textbf{IMU/AHRS }for odometry calculation
	\item Bosch Parkpilot URF7 \textbf{sonar sensors} for collision avoidance (3$\times$)
	\item SEEED ReSpeaker v2.0 \textbf{microphone array} for speech recognition
	\item Velodyne Puck \textbf{3D-Lidar} for localization, people perception, and collision avoidance
	\item Intel D435 \textbf{RGBD cameras} (2$\times$, mounted frontal and dorsal) for localization and people/object perception. The cameras are connected to a Nvidia Jetson Nano inside the head, which provides the synchronized image data via Ethernet and thus reduces the load on the main computer.
\end{itemize}
Apart from the platform, the only actuators are BLDC motors to move the arms and servo motors for the ears.
These are not used for physical interaction or manipulation, but for social interaction to make the appearance of the robot more natural by supporting the emotions to be displayed (e.g. wiggling ears to depict happiness or lowering ears for sadness).
Besides a speaker and tablet (Fig.~\ref{fig:gui}), another main component for interaction is the 64$\times$32 px RGB \textbf{LED matrix} that displays still images or animations.
The speaker, ears and LED panel are controlled by a Raspberry Pi, which communicates with the other computers via Ethernet.

\subsection{Software Architecture}
\begin{figure}[t]
	
	\centering
	
	\includegraphics[width=1\linewidth]{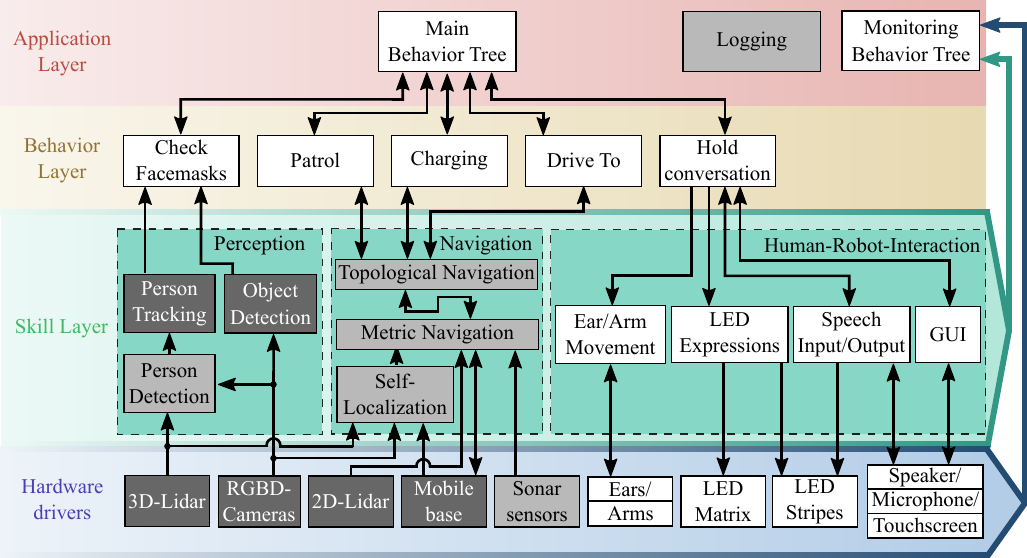}
	
	\caption{Layered software structure of the robotic system. White boxes indicate custom self-developed programs, dark grey boxes third-party ROS programs and light grey boxes modified or extended  third-party ROS programs.}
	
	\label{fig:overview}
	\vspace{-4mm}
\end{figure}

All four computers in Sobi run Linux (Ubuntu) and the \ac{ros} as a framework for communication and control.
The software is partially based on freely available ROS packages, some of which have been adapted for the specific robot setup, but mainly consists of self-developed solutions.
The layered architecture is shown in Fig.~\ref{fig:overview}.
All relevant ROS messages are continuously stored to an external server running a MongoDB instance.
The central control unit in the application layer is a \acf{bt}, partly composed of sub-trees of the behavior layer, which can also be executed independently.
Since the robot  must act purely reactively on immediate requests, modelling via \acp{bt} is well suited and the use of a task scheduler is not necessary.
However, recent advances in \ac{bt} research would also enable more sophisticated use cases which incorporate planning capabilities for \ac{bt} synthesis \cite{Colledanchise2019, Rovida2017}.

In the following sections, the individual software functions of Sobi are presented in more detail.

\subsubsection{People and Object Perception}
People perception and tracking is conducted through the 3D Lidar and RGBD cameras.
YOLO v3 \cite{redmon2018yolov3} is used for people and object detection in the RGB images. The centroids of resulting bounding boxes are registered with a median filter based on the distance either with the 3D Lidar data or the depth data of the cameras.
We use a modified version of \cite{yan2017class} for 3D point cloud segmentation, extended by the approach proposed in \cite{bogoslavskyi2016fast} for 2D clustering to decrease the false negative rate.
The aggregation and tracking is performed with the SPENCER framework \cite{linder2016multi}, which enables tracking in a circular area with a radius of up to \SI{5}{\meter} through the multimodal combination of different sensors.

\subsubsection{Localization and Navigation}
A basic requirement for a mobile robot is to know its own position in relation to the environment.
Especially in dynamically changing environments this poses a challenge. 
As Sobi is designed to be used in different areas, both inside and outside of buildings, a powerful SLAM solution is one of the most important requirements.
RTAB-Map (real-time appearance based mapping)  \cite{Labbe2019} is used as the basic SLAM method, as it allows flexible use of different sensor types (e.g. RGBD cameras, Lidar).
RTAB-Map comprises a powerful framework that can be used for long-term,  large-scale and multi-session mapping.
As input for the SLAM front-end we use the wheel odometry, fused with the measurement data of the IMU via an extended Kalman filter, as odometry source and both cameras and 3D Lidar as sensors.
Since the different types of environment place different demands on the SLAM system, we developed a method that automatically selects predefined SLAM configurations for different environments depending on various criteria, such as distances to walls.
For example, for outdoor or large-scale environments the depth information of the 3D Lidar is used instead of the cameras, and for indoor environments the maximum range for map generation is reduced \cite{Ehlers2020}.

One of the biggest challenges of visual SLAM methods like RTAB-Map are changes in illumination or appearance over time.
To increase the number of loop closures, we added to RTAB-Map the possibility to use 3D point clouds for loop closure detection by a trained classifier of global point cloud descriptors and then register  them in a multi-step process.
%During the use of the robot, it often happened that no loop closure to previously created maps could be found and localization was not possible.
%Therefore, we added to RTAB-Map the possibility to use 3D point clouds for loop closure detection by a trained classifier of global point cloud descriptors and then register  them in a multi-step process.
Especially in outdoor environments and environments poor in visual features, localization could thus be significantly improved.
Further information and experimental results can be found in the corresponding paper \cite{Habich2021}.

For navigation on metric maps, respectively local and global path planning, standard techniques of the ROS framework (i.e. \textit{MoveBase}) are  used.
The 3D map is projected onto the ground plane to create an occupancy grid map that can be used with standard planners.
Above this level, we use navigation on a topological map, where nodes represent either relevant locations, such as specific rooms or facilities, or waypoints between which the robot can either navigate directly or move using problem-specific planners.
%Regular movements between nodes are therefore executed with metric navigation.
%For docking and undocking to the charging station, we developed custom methods.
An example for problem-specific navigation planning is docking and undocking at the charging station.
The former uses a triangular landmark that can be detected by the 2D laser scanner.
It plans a Dubins path consisting of circular- and linear segments and follows the path with a pure pursuit controller \cite{Craig1992}.
These approaches are not complex in their computation and have resulted in robust docking processes in initial evaluations.
\subsubsection{Human-Robot Interaction}
The main task of Sobi is to provide campus-specific information as well as guiding applications. 
For intuitive accessibility, the human-robot interface therefore consists of speech processing as well as touch operation.
Speech processing is based on a combination of Google's Speech-to-Text and Text-to-Speech services and the Natural Language Processing pipeline Dialogflow (see further information in \cite{Stuede2019}).
Since WiFi coverage is available throughout the robot's area of operation, the latencies of the speech processing pipeline are short enough to allow smooth interaction.
In addition, no offline backup solution is required, because the internet connection was available throughout the entire development and test period.
All information can be accessed via both the speech and touch interfaces.
The interface includes the following functions: display and queries of the canteen menu, public transport timetables, staff offices and room locations as well as options for small talk.
By connecting the topological map of the robot with the environmental structure, it is also possible to locate nearby places (e.g. restrooms, seminar rooms or offices).
The path to the requested location can then be displayed on a 3D map for destinations on the same floor (Fig.~\ref{fig:gui}).
A daytime-specific greeting and thus the start of an interaction occurs when a person is detected in front of the robot or the touch display is activated.
To further enhance \ac{hri}, animations on the LED panel, varying colors of the LED strips, and movements of the ears are also used.
Blinking eyes indicated as rings are displayed in normal operation, as well as various animations for specific situations (e.g. \textit{laughing}, \textit{sad} or \textit{sleeping} during the charging process).
When the voice input is activated by a button on the tablet, the robot's LEDs turn green, the ears move forward, and the indicated eyes widen.
Unanswerable requests, on the other hand, are underlined by lowered ears and the depiction of a sad face.

    \section{Monitoring for Long-Term Autonomy}
        	\label{sec:monitoring}
    Monitoring system variables of hardware and software applications is an essential building block for achieving long-term autonomy.
    We therefore developed a framework for Sobi, which monitors various system variables and reacts to faults.
	The monitoring system consists of separate applications that monitor hardware- and ROS framework parameters.
	% and is based on the \textit{monitoring} ROS-package \footnote{\url{https://github.com/luhrts/monitoring}}.
	The hardware monitors include CPU, RAM and network load monitoring as well as measurement of time differences of host computers based on \ac{ntp}.
	ROS nodes are monitored by continuously checking if they are pingable and that essential topic publishing rates  are within a tolerance band.
	Furthermore, we implemented monitors that continuously check whether there is a valid loop closure in the localization system and if there are error cases in the navigation system.
	Each monitor includes an individual warning and error range and respective messages are aggregated on one single \ac{ros} topic.
	The aggregated values are then used as input to an arbiter based on the Behavior Tree framework, that deterministically reacts to the different error cases.
	Although Behavior Trees are mainly used for sequential control of autonomous agents, their advantages in terms of reactivity and modularity combined with intuitive modelability are also applicable for system monitoring.
	Due to the statelessness of BT and their reactive structure, it is thus possible to react immediately to errors that occur, without the need for explicit state transition modeling as in the case of finite state machines, for example.
	The structure of the monitoring system is shown in Fig.~\ref{fig:mon_bt}.

	The system entities (i.e. programs/topics) to monitor are organized in configurations, that can be switched autonomously or manually and are created for different use cases, e.g. for normal operation, charging or mapping.
	All unneeded programs from other configurations are terminated when a configuration change is made.
	Whether an error is present is determined on a monitor-specific basis via detection signals.
	Table~\ref{tab:monitors} summarizes these signals together with the associated recovery reaction.

	Similar to \cite{Meeussen2011}, we follow the approach that maximum robustness may not be achieved by full autonomy alone, but by planned interventions of supervisors in case of failure. 
	Therefore, the last resort for navigation and localization errors is a predefined request to a list of supervisors, in which the robot sends an instant message with an URL.
	On the linked website, a teleoperation can then be performed based on the camera views, or the current position on the map can be specified.
	This simple system intervention usually takes less than a minute of the operator's time, but prevents a total failure of the system.
	Once the problem has been fixed, this is confirmed by the operator and an all-clear is sent to the other supervisors.
	
   %\begin{table}
   	%\caption{Overview of the utilized monitors with detection signal and corresponding reaction}
   	%\input{monitors_table.tex}
   %\end{table}
\begin{figure}[!t]
    		\noindent
\begin{minipage}[t]{\linewidth}
	%	\centering
	
	%	\begin{minipage}[h]{\linewidth}
		\begin{minipage}[]{\linewidth}
		\centering
		\resizebox{\columnwidth}{!}{%
				\begin{tikzpicture}
		   [Monitor/.style = {rectangle, draw, align=center, 
			inner ysep=3mm, minimum width = 23mm, fill=white, minimum height=6mm},
		MonWide/.style = {rectangle, draw, align=center, 
			inner ysep=2mm, minimum width = 35mm, fill=white, minimum height=8mm},
	Agg/.style = {rectangle, draw, align=center, 
			inner ysep=0mm, minimum width = 6mm, minimum height=2mm, fill=white},
			CfgGroup/.style={rectangle, draw, minimum width=2.8cm,
				minimum height = 5.5cm},
			GroupLabel/.style={minimum width=0.8cm,minimum height = 1.0cm, text width=1cm, align=center},
		-{Latex[length=3mm,width=2mm]},
		]

		\node[Monitor] (NodeMon) {Node Mon.};
		\node[Monitor, below = 2mm of NodeMon.south] (TopicMon) {Topic Mon.};
		\node[Monitor, below = 2mm of TopicMon.south] (NavMon) {Navigation\\ Mon.};
		\node[Monitor, below = 2mm of NavMon.south] (LocMon) {Localization\\ Mon.};
	  \begin{pgfonlayer}{c3}
		\node[CfgGroup, fit={(NodeMon) (TopicMon) (NavMon) (LocMon)}, draw, label={[shift=(GroupOperating.north west), align=left, anchor=south west]Configurations\\ \tikzmarknode[,  top color=gray!10, bottom color=gray!30,draw=black,thick,inner sep=2pt,text opacity=1]{Ch}{\textit{Operating}},		\tikzmarknode[,  top color=red!10, bottom color=red!30,draw=black,thick,inner sep=2pt,text opacity=1]{Ch}{\textit{Charging}}, \tikzmarknode[,  top color=green!10, bottom color=green!30,draw=black,thick,inner sep=2pt,text opacity=1]{Ch}{\textit{Mapping}}, ...}, ,  top color=gray!10, bottom color=gray!30] (GroupOperating) {};

		%\node[GroupLabel, above left = 5mm and 3mm of GroupOperating.north west, anchor=south] (OperatingLabel) {Config \textit{Operating}};
	%	\draw[-,  bend angle=15, bend right] (OperatingLabel) to (GroupOperating.north west);	
		\end{pgfonlayer}
	\begin{pgfonlayer}{c2}
		\node[CfgGroup, below left =2mm and 2mm of GroupOperating.north east, label={}, ,  top color=red!10, bottom color=red!30] (GroupCharging) {};
	%	\node[GroupLabel, below left = 0mm and 8mm of GroupCharging.north west, anchor=north east] (ChargingLabel) {Config\\ \textit{Charging}};
	%	\draw[-,  bend angle=15, bend left] (ChargingLabel.north) to (GroupCharging.north west);
	\end{pgfonlayer}
		\begin{pgfonlayer}{c1}
		\node[CfgGroup, below left =2mm and 2mm of GroupCharging.north east, label={}, ,  top color=green!10, bottom color=green!30] (GroupMapping) {};
	%	\node[GroupLabel, below left = 29mm and 6mm of GroupMapping.north west, anchor=north east] (MappingLabel) {Config\\ \textit{Mapping}};
	%	\draw[-,  bend angle=15, bend left] (MappingLabel.north) to (GroupMapping.west);
	\end{pgfonlayer}
		\begin{pgfonlayer}{bg}
	\node[CfgGroup, below left =3mm and 3mm of GroupMapping.north east, fill=white, align=left] (GroupRest) {};
			\node[above right = 0.5mm and 0.5mm of GroupRest.south west, anchor=south west] (MappingDots) {...};
\end{pgfonlayer}
		%		\node[ellipse,label=270:OT cl,align=left,draw] (a) at (0,0) {Simulation\\ Objects};
		\node[MonWide, below=8mm of GroupOperating.south] (ChangeCfg) {\subnode{sCfg}{}Change Configuration};
			\node[Agg, label={[label distance=0.0cm,text depth=0ex,rotate=90]center:Aggregation}, right=20mm of GroupOperating.east, anchor=west, fit=(NodeMon.north) (LocMon.south), inner sep=0] (Aggr) {\subnode{sAgg}{}};

		 \draw (ChangeCfg.north) to (GroupOperating.south);
		\draw (NodeMon) -- (NodeMon-|Aggr.west);
		\draw (TopicMon) -- (TopicMon-|Aggr.west);
		\draw (NavMon) -- (NavMon-|Aggr.west);
		\draw (LocMon) -- (LocMon-|Aggr.west);
		\node[rectangle,align=left,draw, label={Monitoring Behavior Tree}, right = 12 mm of GroupOperating.north east, anchor=north west](BT){
				\begin{forest}
				for tree={%
					l sep=0.6cm,
					s sep=0.1cm,
					edge={->, >=latex},
					parent anchor=south,
					child anchor=north,
					align=center,
					anchor=center,
					%drop shadow
				}
				[, bt-root
				[, bt-sequence
				[\subnode{sBtConf}{}Check \& Set\\Configuration, name=SetConfiguration, bt-subtree]
				[\subnode{sBtProc}{}Process\\Monitors, bt-subtree]{\draw[->,dashed,>=latex, line width=0.7pt] ()  to[out=east,in=west] (Monitor-Fallback);}
				[,name=Monitor-Fallback, bt-fallback
				%	[Customer] {\draw[->,dotted] () -- ++(-4cm,0) to[out=north,in=west] (Company);}
				[,bt-sequence
				[, bt-fallback
				[Node\\error, bt-condition]
				[Topic\\error, bt-condition]
				]
				[Restart\\Node, bt-subtree]
				]
				[,bt-sequence
				[Nav.\\error, bt-condition]
				[Recover\\Navigation, bt-subtree]
				]
				[,bt-sequence
				[Loc.\\error, bt-condition]
				[Recover\\Localization, bt-subtree]
				]
				[, bt-etc]
				]
				]
				]	
			\end{forest}

	};
	\node[MonWide, inner sep=0, anchor=south] (SysMons) at (ChangeCfg |- BT.south) {System Monitors};
	\node[, inner sep=0, anchor=north east, below left = 2mm and 5mm of BT.north east, minimum height=3mm] (legendText) {Legend:};
	\node[bt-condition, inner sep=0, anchor=north, below = 1mm of legendText, minimum height=8mm] (legendCond) {Condition};
   \node[bt-subtree, inner sep=0, anchor=north, below = 1mm of legendCond, minimum height=8mm, minimum width=13mm] (legendSubtree) {Subtree};
	 \draw (SysMons) -| (Aggr);
	%	\draw[latex-latex] (a) -- (b);
	%	\node (c) at ($(a.south west)+(-1,-1)$) {};
	%\node (d) at ($(b.north east)+(1,1)$) {};
	%	\node[fit=(a)(d),draw] (e) {};
	%	\draw ($(e.south west)+(\pgflinewidth/2,0)$) rectangle node {NS2 Shell Executable Command (ns)} ($(e.south east)+(-\pgflinewidth/2,-0.7)$);
	\end{tikzpicture}
	% This arrow should connect two nodes of both trees.
	\begin{tikzpicture}[overlay,remember picture]
			\draw[->, black, dashed,-{Latex[length=3mm,width=2mm]}, line width=1.2pt, bend angle=25, bend left] ($(pic cs:sAgg)+(0.3,.0)$) to ($(pic cs:sBtProc)+(0.3,0.45)$);	
			\draw[->, black, dashed,-{Latex[length=3mm,width=2mm]}, line width=1.2pt, bend angle=25, bend left] ($(pic cs:sBtConf)+(-0.2,-.7)$) to ($(pic cs:sCfg)+(3.35,0.0)$);
			
	%	\draw[->, black, dashed, bend angle=45, bend left] ($(pic cs:replaceNode)+(0.25,.2)$) to ($(pic cs:t1)+(0.25,.2)$);
	\end{tikzpicture}
		}
		\captionof{figure}{Structure of the monitoring framework. Configuration dependent monitored values are processed in a \ac{bt}. The composition of the tree defines the priority of error handling.}
		
		\label{fig:mon_bt}
		%\vspace{-7mm}
	\end{minipage}
	% \hfill
	\begin{minipage}[]{\linewidth}
		\vspace{3mm}
		%	\vspace{-2mm}
		%	\centering
		\captionsetup{font=small}
		%	\captionof{table}{Overview of the utilized monitors with detection signal and corresponding reaction}
		\captionof{table}{Overview of the utilized monitors with detection signal and corresponding reaction.}
		\begin{adjustbox}{width=\linewidth,center}
			
			\def\arraystretch{1.1}
\begin{tabular}{l|p{3.6cm}|p{3.4cm}}
	%\rule{0pt}{0ex}

	\textbf{Name }& \textbf{Detection} & \textbf{Reaction}\\
\hline
	Node monitor & Node not pingable & Restart node \\
	Topic monitor & Publish frequency not in tolerance band & Restart publishing node\\[3mm]
	\makecell[l]{CPU, RAM, NTP, Network \\ Monitor} & Not in tolerance band & Send message \\[3mm]
	Navigation monitor & No global path found and \textit{MoveBase} recoveries failed &
	\begin{minipage}[t]{\linewidth}
	\begin{enumerate}[leftmargin=0.4cm,after=\vspace{-\baselineskip},before=\vspace{-0.5\baselineskip}]
		\item Wait and retry
		\item Move backwards
		\item Ask Supervisor
	\end{enumerate}
	\end{minipage} \\[10mm]

Localization monitor & No loop closure detected &
	\begin{minipage}[t]{\linewidth}
	\begin{enumerate}[leftmargin=0.4cm,after=\vspace{-\baselineskip},before=\vspace{-0.5\baselineskip}]
	\item Slow rotate
	\item Restart localization
	\item Slow rotate
	\item Ask Supervisor
\end{enumerate}
%\vspace{-2mm}
\end{minipage}\\
	
\end{tabular}
		\end{adjustbox}
		
		\label{tab:monitors}
	%	\vspace{5mm}
	\end{minipage}

\end{minipage}
\end{figure}

    \section{Evaluation}
    	\label{sec:evaluation}

    	\begin{figure*}[t]
    		    		\begin{minipage}[]{0.5\textwidth}
    			
    			\centering
    			\includegraphics[width=1\linewidth]{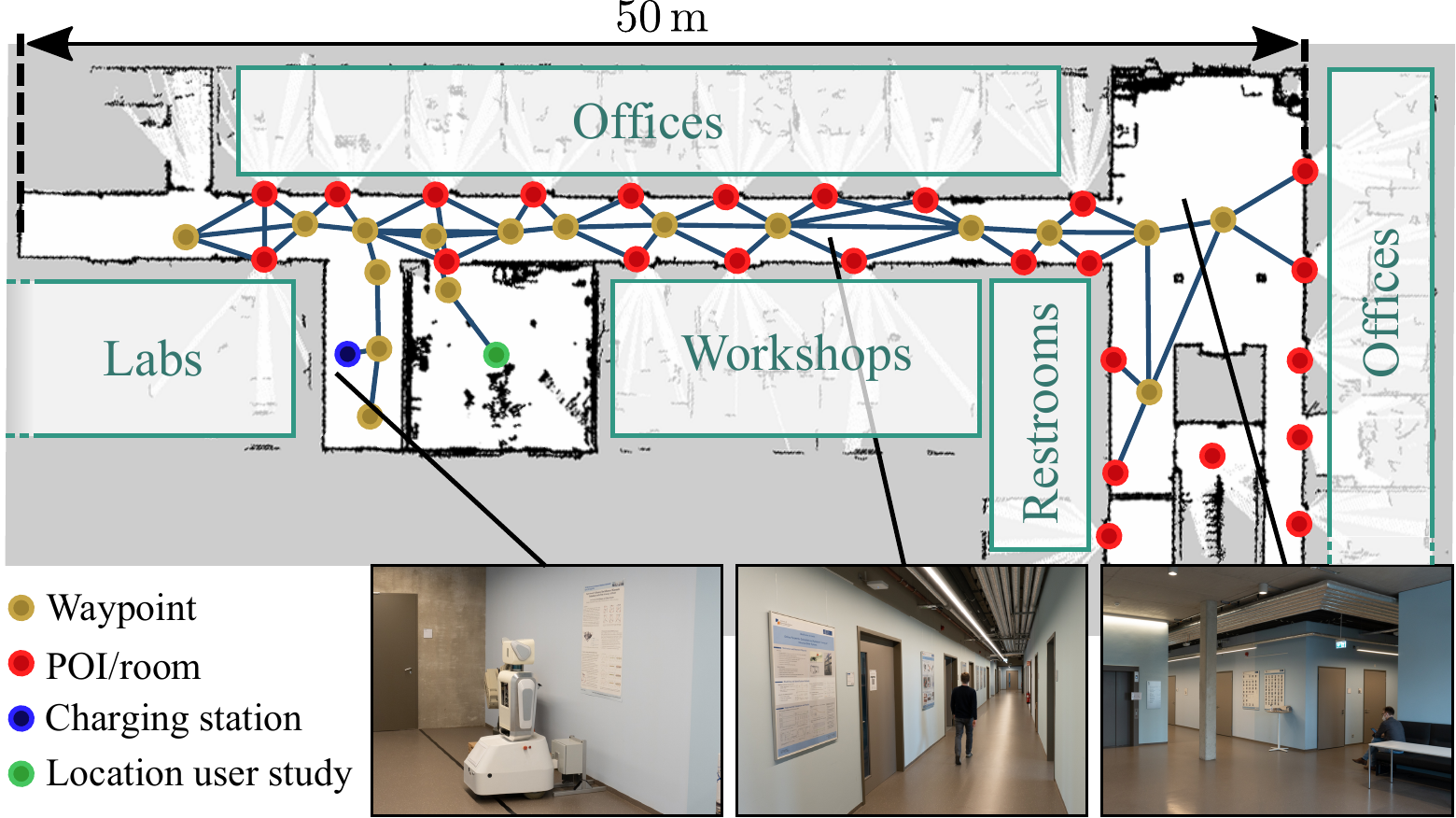}
    			\captionof{figure}{Environment used for validation. Dots indicate nodes and blue lines indicate edges of the topological map. The robot continuously patrols between the shown waypoints in a random fashion.}
    			
    			\label{fig:environment}
    			
    		\end{minipage}
    		\hfill
    		\begin{minipage}[]{0.5\textwidth}
    			%\vspace{5mm}
    			\hfill
\begin{minipage}[t]{.42\linewidth}
	\def\arraystretch{1.1}
	
	\captionsetup{font=small}
	\captionof{table}{Metrics of the deployment. Weekend days are excluded.}
	
	\label{tab:lta}
	\begin{adjustbox}{width=1\linewidth,center}
		\begin{tabular}{p{2.8cm}|p{1.3cm}}
			%\rule{0pt}{0ex}
			
			\textbf{Metric}&\textbf{Value}\\
			\hline
			Timespan& 12 days\\
			\hline
			Mean time operating per day& \SI{5.9}{\hour}\\
			\hline
			Mean time moving per day $t_\mathrm{m}$&\SI{5.5}{\hour}\\
			\hline
			$A\%=\nicefrac{t_\mathrm{m}}{\SI{8}{\hour}}$&\SI{69}{\percent}\\
			%	\hline
			%	Time charging&\SI{312.7}{\hour}\\
			\hline
			Traveled distance&\SI{66.6}{\kilo\meter}\\
			\hline
			Mean TSL&\SI{56.7}{\hour}\\
			\hline
			Max TSL&\SI{90.6}{\hour}\\
			\hline
			Detected faces with mask&983\\
			\hline
			Detected faces w/o mask&101\\
			\hline
			%	Docking attempts&28\\
			%	\hline
		\end{tabular}
	\end{adjustbox}
\end{minipage}
\hfill
\begin{minipage}[t]{.5\linewidth}
	\def\arraystretch{1.3}
	
	\captionsetup{font=small}
	\captionof{table}{Executed recovery behaviors sorted by category.}
	
	\label{tab:eval}
	\vspace{2.3mm}
	\begin{adjustbox}{width=1\linewidth,center}
		
		\begin{tabular}{p{0.5cm}  p{2.7cm}  p{0.3cm} p{0.9cm}}
			%\rule{0pt}{0ex}
			
			\textbf{Type}&\textbf{Reaction}&\textbf{\#}&\textbf{success}\\
			\cmidrule{1-4}
			ROS&Restart node&32& \SI{62.5}{\percent}\\\addlinespace[0.3cm]
			\multirow{3}{*}[2.5ex]{\ldelim\{{3}{0.6cm}[{\rotatebox[origin=c]{90}{\ Navigation\ }}]}&Wait and retry&194& \SI{70.6}{\percent}\\
			&Move backwards&41&\SI{48.8}{\percent}\\
			&Ask supervisor for teleoperation&5&\SI{100}{\percent}\\\addlinespace[0.3cm]
			%\ldelim\{{4}{*}[\rotatebox{90}{Localization}]&Slow rotate&76&\SI{84.2}{\percent}\\
			\multirow{3}{*}[2.5ex]{\ldelim\{{4}{0.5cm}[{\rotatebox[origin=c]{90}{\ Localization\ }}]}&Slow rotate&76&\SI{84.2}{\percent}\\
			&Restart localiza- tion \& rotate again&12&\SI{66.6}{\percent}\\
			&Ask supervisor for relocalization&4&\SI{100}{\percent}\\
			
		\end{tabular}
		
	\end{adjustbox}
\end{minipage}

    		\end{minipage}
    		\vspace{-3mm}
    	\end{figure*}
    The evaluation of the robot consists of two parts: In the first part, we test the \ac{lta} capabilities in a continuous 16 day deployment and in the second part, we conduct a user study on human-robot interaction to evaluate the user interface and robot's appearance.
    	\subsection{Long-Term Autonomy}
    The goal of this evaluation is to determine the robustness of the localization, navigation and monitoring system as well as the individually created algorithms, e.g. for docking, and to identify possible weaknesses.
    The robot's task is to permanently patrol within one floor during office hours (9 am--5 pm) on working days.
    Outside these hours and if necessary in between, Sobi autonomously approaches the charging station and performs the charging process.
    The environment mainly consists of a hallway with several offices and an entrance area.
    The metric and topological maps of the environment are shown in Fig.~\ref{fig:environment}.
    Due to hygiene restrictions in place at the time of the evaluation, passers-by are not allowed to access the robot by touch  and are also required to wear a face mask while in the building.
    The robot is therefore tasked with detecting whether people are wearing a face mask, by analyzing the image of the frontal camera every \SI{0.5}{\second} and using face detection.
    The underlying YOLO network was trained with 600 images containing both \textit{mask} and \textit{no mask} classes.
    A test set of 100 images results in an average precision of \SI{92.9}{\percent} for the \textit{mask} and  \SI{82.34}{\percent} for the \textit{no mask} class.
    If the \textit{no mask} class is detected, the robot verbally asks the person to put on a mask.
    Given the total number of detections in the evaluation period, the average precision can give a rough estimate about how many people were encountered directly in front of the robot.
    This therefore also serves as a measure of the environment dynamics, since crowded environments pose a greater challenge for long-term autonomous systems.
   	Together with typical metrics for long-term autonomy the number of detections is summarized in Table \ref{tab:lta}.
	In total, Sobi was undocked from the charging station for \SI{71.2}{\hour}, of which \SI{65.4}{\hour} were spent in motion, so that a distance of \SI{66.6}{\kilo\meter} was covered during the evaluation period.
	The time in motion indicates the time during which the robot was able to perform the patrolling task and was not in an error or recovery state.
	With respect to an 8-hour workday of the robot, this results in an autonomy percentage ($A\%$) of \SI{69}{\percent}, which is an indicator of the percentage of the available time that the robot actually uses to perform its services (patrolling).
	In our case, this indicative parameter is mainly influenced by the average additional charging time of \SI{2}{\hour} during the day.
	
	An overview of the recovery behaviors performed is shown in Table \ref{tab:eval}. As in \cite{Hawes2016}, a recovery behavior is considered successful if no further recovery behavior needs to be performed within one minute.
	For the recovery behaviors for navigation, a further requirement is that no additional recovery behavior must be performed within a circle with a radius of \SI{1}{\meter} within this time.
	From the table it can be seen that many problems can already be solved by waiting and going back (in case of navigation errors) or slow rotation and restarts (in case of localization errors).
	Requested remote access was necessary in a total of nine cases. These occurred more frequently when there were major changes to the environment and, for example, many doors were open that had been closed during the mapping process, or when closed fire doors or obstacles blocked the way.
	
	A typical quantity for evaluating robustness is the \ac{tsl}, which specifies the time interval between interventions by supervisors in the event of a failure, which were not specifically requested by the robot.
	During the evaluation period, a total of four of such interventions were necessary, resulting in the  \ac{tsl} values shown.
	Two of the interventions were due to software errors that resulted in the signal to approach the charging station not being sent, requiring manual intervention to avoid shutdown. 
	On one other occasion, various programs were permanently restarted within a deadlock state, making it impossible to continue the process.
	This deadlock situation mainly influences the \SI{62.5}{\percent} success rate of the node restart recovery, because ten node restarts occurred in a timespan of less than three minutes.
	The fourth intervention was necessary when a docking attempt failed, so that the robot performed an emergency shutdown and had to be switched on again manually.
	A total of 28 docking attempts were made during the evaluation period, with all other 27 attempts carried out successfully.
	Except for this last error, all problems could be recovered via remote SSH access to the system and no hardware errors occurred.
	One drawback of the \ac{tsl}, and also the $A\%$ is, that it does not validate how \emph{useful} the services of the robot actually are for users.
	In the next section, we therefore present a separate study on how users perceive the robot and its info-terminal services.

% Quellen:
% \cite{Bartneck2009}: DOI 10.1007/s12369–008–0001–3
% \cite{Weiss2015}: DOI 10.1109/ROMAN.2015.7333568
% [27]: The measurement of meaning, 1957 (Review: DOI 10.1017/S0008413100018740)
% [28]: DOI 10.1007/BF02310555

\subsection{User Study on Human-Robot Interaction}
The user study is based on the \ac{gqs}~\cite{Bartneck2009} and is conducted with 12 participants (10 male, 2 female, all between 25 and 34 years old), who had no prior knowledge or direct contact with the robot before the study. 
The \ac{gqs} measures a user's perceived sensations during a social interaction with a robot and is one of the most widely used assessment criteria in this area~\cite{Weiss2015}.
It describes the perceived impression of the robot via five categories and different bipolar adjective pairs on a scale from 1 to 5 using a semantic differential.% according to Osgood~[27].

%With the standardized evaluation criteria of the GQS, the impression of the robot perceived during an interaction can be described holistically 
%Within each category, different bipolar adjective pairs are considered, which are evaluated on the basis of emotional decisions on a scale from 1 to 5 using a semantic differential.% according to Osgood~[27].

The interaction between Sobi and a user is conducted in the location shown in Fig.~\ref{fig:environment} and is divided into three independent phases.
During the experiment, the robot remained in a static position and interaction was started by the users.
The shown POI are visualized on the 3D map of the tablet accordingly.
In the first phase, users were given specific tasks to perform. 
This included figuring out how to get to the nearest restroom as well as to the office of a given person. 
Users were not provided with any assistance, nor were they made aware that there was more than one possibility to get the requested information, using speech commands or the tablet GUI. 
After successfully visualizing the path to the destination on the tablet's 3D map, users should identify information about public transportation departure times. 
In the second phase of the study, five minutes were provided for free interaction, during which Sobi's abilities were to be learned in more detail by trying them out for themselves. 
In a concluding observation phase, Sobi gave some verbal closing remarks and insight about further  possible emotions and robot states. 

\begin{figure*}[t]
	\setlength\figureheight{2.0cm}
	
	\setlength\sfwidth{0.24\textwidth}
	\setlength\figurewidth{\sfwidth}
	\definecolor{none}{rgb}{0,0,0}
	\captionsetup[subfigure]{justification=centering}
	\centering
	%    \subcaptionbox{%
	%    	Anthropomorphism \\ ($\alpha=0.84$)
	%    }[0.833\sfwidth]
	%    {\input{antropomorph.tex}}
	%    \hfill
	%        \subcaptionbox{%
	%    	Likeability \\ ($\alpha=0.83$)
	%    }[0.833\sfwidth]
	%    {\input{likeability.tex}}
	%        \hfill
	%    \subcaptionbox{%
	%    	Animacy \\ ($\alpha=0.66$)
	%    }[\sfwidth]
	%    {\input{animacy.tex}}
	%            \hfill
	%    \subcaptionbox{%
	%    	Perceived Intelligence \\ ($\alpha=0.90$)
	%    }[0.833\sfwidth]
	%    {\input{intelligence.tex}}
	%                \hfill
	%    \subcaptionbox{%
	%    	Perceived Safety \\ ($\alpha=-0.20$)
	%    }[0.5\sfwidth]
	%    {\input{safety.tex}}
	\begin{subfigure}[t]{0.833\sfwidth}
		%\hspace{1mm}
		\setlength\figurewidth{\textwidth-2mm}
		\hspace*{1mm}
\begin{tikzpicture}[%
baseline=1cm, trim axis left, trim axis right
]

\begin{axis}[%
width=0.951\figurewidth,
height=\figureheight,
at={(0\figurewidth,0\figureheight)},
scale only axis,
unbounded coords=jump,
separate axis lines,
every outer x axis line/.append style={black},
every x tick label/.append style={font=\color{black}},
every x tick/.append style={black},
xmin=0.5,
xmax=5.5,
xtick={1,2,3,4,5},
xticklabels={{fake},{machinelike},{unconscious},{moving rigidly},{artificial}},
every outer y axis line/.append style={black},
every y tick label/.append style={font=\color{black}},
every y tick/.append style={black},
ymin=0.85,
ymax=5.15,
ytick={1, 2, 3, 4, 5},
axis background/.style={fill=white},
ymajorgrids,
xticklabel style={rotate=30, anchor=east, font=\scriptsize, yshift=-0.1cm, xshift=0.05cm}, yticklabel style={font=\scriptsize}
]
\addplot [color=black, dashed, forget plot]
  table[row sep=crcr]{%
1	4\\
1	5\\
};
\addplot [color=black, dashed, forget plot]
  table[row sep=crcr]{%
2	3\\
2	3\\
};
\addplot [color=black, dashed, forget plot]
  table[row sep=crcr]{%
3	4\\
3	4\\
};
\addplot [color=black, dashed, forget plot]
  table[row sep=crcr]{%
4	4\\
4	5\\
};
\addplot [color=black, dashed, forget plot]
  table[row sep=crcr]{%
5	3.5\\
5	4\\
};
\addplot [color=black, dashed, forget plot]
  table[row sep=crcr]{%
1	2\\
1	2\\
};
\addplot [color=black, dashed, forget plot]
  table[row sep=crcr]{%
2	1\\
2	2\\
};
\addplot [color=black, dashed, forget plot]
  table[row sep=crcr]{%
3	1\\
3	2\\
};
\addplot [color=black, dashed, forget plot]
  table[row sep=crcr]{%
4	2\\
4	2.5\\
};
\addplot [color=black, dashed, forget plot]
  table[row sep=crcr]{%
5	2\\
5	2\\
};
\addplot [color=black, forget plot]
  table[row sep=crcr]{%
0.875	5\\
1.125	5\\
};
\addplot [color=black, forget plot]
  table[row sep=crcr]{%
1.875	3\\
2.125	3\\
};
\addplot [color=black, forget plot]
  table[row sep=crcr]{%
2.875	4\\
3.125	4\\
};
\addplot [color=black, forget plot]
  table[row sep=crcr]{%
3.875	5\\
4.125	5\\
};
\addplot [color=black, forget plot]
  table[row sep=crcr]{%
4.875	4\\
5.125	4\\
};
\addplot [color=black, forget plot]
  table[row sep=crcr]{%
0.875	2\\
1.125	2\\
};
\addplot [color=black, forget plot]
  table[row sep=crcr]{%
1.875	1\\
2.125	1\\
};
\addplot [color=black, forget plot]
  table[row sep=crcr]{%
2.875	1\\
3.125	1\\
};
\addplot [color=black, forget plot]
  table[row sep=crcr]{%
3.875	2\\
4.125	2\\
};
\addplot [color=black, forget plot]
  table[row sep=crcr]{%
4.875	2\\
5.125	2\\
};
\addplot [color=blue, forget plot]
  table[row sep=crcr]{%
0.75	2\\
0.75	4\\
1.25	4\\
1.25	2\\
0.75	2\\
};
\addplot [color=blue, forget plot]
  table[row sep=crcr]{%
1.75	2\\
1.75	3\\
2.25	3\\
2.25	2\\
1.75	2\\
};
\addplot [color=blue, forget plot]
  table[row sep=crcr]{%
2.75	2\\
2.75	4\\
3.25	4\\
3.25	2\\
2.75	2\\
};
\addplot [color=blue, forget plot]
  table[row sep=crcr]{%
3.75	2.5\\
3.75	4\\
4.25	4\\
4.25	2.5\\
3.75	2.5\\
};
\addplot [color=blue, forget plot]
  table[row sep=crcr]{%
4.75	2\\
4.75	3.5\\
5.25	3.5\\
5.25	2\\
4.75	2\\
};
\addplot [color=red, line width=1.5pt, forget plot]
  table[row sep=crcr]{%
0.75	3.5\\
1.25	3.5\\
};
\addplot [color=red, line width=1.5pt, forget plot]
  table[row sep=crcr]{%
1.75	2.5\\
2.25	2.5\\
};
\addplot [color=red, line width=1.5pt, forget plot]
  table[row sep=crcr]{%
2.75	3\\
3.25	3\\
};
\addplot [color=red, line width=1.5pt, forget plot]
  table[row sep=crcr]{%
3.75	3\\
4.25	3\\
};
\addplot [color=red, line width=1.5pt, forget plot]
  table[row sep=crcr]{%
4.75	3\\
5.25	3\\
};
\addplot [color=black, only marks, mark=+, mark options={solid, draw=blue}, forget plot]
  table[row sep=crcr]{%
nan	nan\\
};
\addplot [color=black, only marks, mark=+, mark options={solid, draw=blue}, forget plot]
  table[row sep=crcr]{%
2	5\\
};
\addplot [color=black, only marks, mark=+, mark options={solid, draw=blue}, forget plot]
  table[row sep=crcr]{%
nan	nan\\
};
\addplot [color=black, only marks, mark=+, mark options={solid, draw=blue}, forget plot]
  table[row sep=crcr]{%
nan	nan\\
};
\addplot [color=black, only marks, mark=+, mark options={solid, draw=blue}, forget plot]
  table[row sep=crcr]{%
nan	nan\\
};
\end{axis}

\begin{axis}[%
width=0.951\figurewidth,
height=\figureheight,
at={(0\figurewidth,0\figureheight)},
scale only axis,
every outer x axis line/.append style={black},
every x tick label/.append style={font=\color{black}},
every x tick/.append style={black},
xmin=0.5,
xmax=5.5,
xtick={1,2,3,4,5},
xticklabels={{natural},{humanlike},{conscious},{moving elegantly},{lifelike}},
every outer y axis line/.append style={none},
every y tick label/.append style={font=\color{none}},
every y tick/.append style={none},
ymin=0.85,
ymax=5.15,
ytick={1,1.5,2,2.5,3,3.5,4,4.5,5},
yticklabels={\empty},
axis x line*=top,
axis y line*=left,
xticklabel style={rotate=30, anchor=west, font=\scriptsize, yshift=0.15cm, xshift=-0.1cm}, yticklabel style={font=\scriptsize}
]
\end{axis}
\end{tikzpicture}%
		%	\caption{Anthropomorphism \\ ($\alpha=0.84$)}
	\end{subfigure}
	\hfill
	\begin{subfigure}[t]{0.833\sfwidth}
		\setlength\figurewidth{\textwidth}
\begin{tikzpicture}[%
baseline=1cm, trim axis left, trim axis right
]

\begin{axis}[%
width=0.951\figurewidth,
height=\figureheight,
at={(0\figurewidth,0\figureheight)},
scale only axis,
unbounded coords=jump,
separate axis lines,
every outer x axis line/.append style={black},
every x tick label/.append style={font=\color{black}},
every x tick/.append style={black},
xmin=0.5,
xmax=5.5,
xtick={1,2,3,4,5},
xticklabels={{dislike},{unfriendly},{unkind},{unpleasant},{awful}},
every outer y axis line/.append style={black},
every y tick label/.append style={font=\color{black}},
every y tick/.append style={black},
ymin=0.85,
ymax=5.15,
ytick={1,2,3,4,5},
yticklabels={\empty},
axis background/.style={fill=white},
ymajorgrids,
xticklabel style={rotate=30, anchor=east, font=\scriptsize, yshift=-0.1cm, xshift=0.05cm}, yticklabel style={font=\scriptsize}
]
\addplot [color=black, dashed, forget plot]
  table[row sep=crcr]{%
1	5\\
1	5\\
};
\addplot [color=black, dashed, forget plot]
  table[row sep=crcr]{%
2	5\\
2	5\\
};
\addplot [color=black, dashed, forget plot]
  table[row sep=crcr]{%
3	5\\
3	5\\
};
\addplot [color=black, dashed, forget plot]
  table[row sep=crcr]{%
4	5\\
4	5\\
};
\addplot [color=black, dashed, forget plot]
  table[row sep=crcr]{%
5	5\\
5	5\\
};
\addplot [color=black, dashed, forget plot]
  table[row sep=crcr]{%
1	4\\
1	4\\
};
\addplot [color=black, dashed, forget plot]
  table[row sep=crcr]{%
2	5\\
2	5\\
};
\addplot [color=black, dashed, forget plot]
  table[row sep=crcr]{%
3	4\\
3	4\\
};
\addplot [color=black, dashed, forget plot]
  table[row sep=crcr]{%
4	4\\
4	4.5\\
};
\addplot [color=black, dashed, forget plot]
  table[row sep=crcr]{%
5	4\\
5	4\\
};
\addplot [color=black, forget plot]
  table[row sep=crcr]{%
0.875	5\\
1.125	5\\
};
\addplot [color=black, forget plot]
  table[row sep=crcr]{%
1.875	5\\
2.125	5\\
};
\addplot [color=black, forget plot]
  table[row sep=crcr]{%
2.875	5\\
3.125	5\\
};
\addplot [color=black, forget plot]
  table[row sep=crcr]{%
3.875	5\\
4.125	5\\
};
\addplot [color=black, forget plot]
  table[row sep=crcr]{%
4.875	5\\
5.125	5\\
};
\addplot [color=black, forget plot]
  table[row sep=crcr]{%
0.875	4\\
1.125	4\\
};
\addplot [color=black, forget plot]
  table[row sep=crcr]{%
1.875	5\\
2.125	5\\
};
\addplot [color=black, forget plot]
  table[row sep=crcr]{%
2.875	4\\
3.125	4\\
};
\addplot [color=black, forget plot]
  table[row sep=crcr]{%
3.875	4\\
4.125	4\\
};
\addplot [color=black, forget plot]
  table[row sep=crcr]{%
4.875	4\\
5.125	4\\
};
\addplot [color=blue, forget plot]
  table[row sep=crcr]{%
0.75	4\\
0.75	5\\
1.25	5\\
1.25	4\\
0.75	4\\
};
\addplot [color=blue, forget plot]
  table[row sep=crcr]{%
1.75	5\\
1.75	5\\
2.25	5\\
2.25	5\\
1.75	5\\
};
\addplot [color=blue, forget plot]
  table[row sep=crcr]{%
2.75	4\\
2.75	5\\
3.25	5\\
3.25	4\\
2.75	4\\
};
\addplot [color=blue, forget plot]
  table[row sep=crcr]{%
3.75	4.5\\
3.75	5\\
4.25	5\\
4.25	4.5\\
3.75	4.5\\
};
\addplot [color=blue, forget plot]
  table[row sep=crcr]{%
4.75	4\\
4.75	5\\
5.25	5\\
5.25	4\\
4.75	4\\
};
\addplot [color=red, line width=1.5pt, forget plot]
  table[row sep=crcr]{%
0.75	5\\
1.25	5\\
};
\addplot [color=red, line width=1.5pt, forget plot]
  table[row sep=crcr]{%
1.75	5\\
2.25	5\\
};
\addplot [color=red, line width=1.5pt, forget plot]
  table[row sep=crcr]{%
2.75	5\\
3.25	5\\
};
\addplot [color=red, line width=1.5pt, forget plot]
  table[row sep=crcr]{%
3.75	5\\
4.25	5\\
};
\addplot [color=red, line width=1.5pt, forget plot]
  table[row sep=crcr]{%
4.75	5\\
5.25	5\\
};
\addplot [color=black, only marks, mark=+, mark options={solid, draw=blue}, forget plot]
  table[row sep=crcr]{%
nan	nan\\
};
\addplot [color=black, only marks, mark=+, mark options={solid, draw=blue}, forget plot]
  table[row sep=crcr]{%
2	4\\
2	4\\
};
\addplot [color=black, only marks, mark=+, mark options={solid, draw=blue}, forget plot]
  table[row sep=crcr]{%
nan	nan\\
};
\addplot [color=black, only marks, mark=+, mark options={solid, draw=blue}, forget plot]
  table[row sep=crcr]{%
4	3\\
};
\addplot [color=black, only marks, mark=+, mark options={solid, draw=blue}, forget plot]
  table[row sep=crcr]{%
nan	nan\\
};
\end{axis}

\begin{axis}[%
width=0.951\figurewidth,
height=\figureheight,
at={(0\figurewidth,0\figureheight)},
scale only axis,
every outer x axis line/.append style={black},
every x tick label/.append style={font=\color{black}},
every x tick/.append style={black},
xmin=0.5,
xmax=5.5,
xtick={1,2,3,4,5},
xticklabels={{like},{friendly},{kind},{pleasant},{nice}},
every outer y axis line/.append style={none},
every y tick label/.append style={font=\color{none}},
every y tick/.append style={none},
ymin=0.85,
ymax=5.15,
ytick={1,1.5,2,2.5,3,3.5,4,4.5,5},
yticklabels={\empty},
axis x line*=top,
axis y line*=left,
xticklabel style={rotate=30, anchor=west, font=\scriptsize, yshift=0.15cm, xshift=-0.1cm}, yticklabel style={font=\scriptsize}
]
\end{axis}
\end{tikzpicture}%
		%	\caption{Likeability \\ ($\alpha=0.83$)}
	\end{subfigure}
	\hfill
	\begin{subfigure}[t]{\sfwidth}
		\setlength\figurewidth{\textwidth}
\begin{tikzpicture}[%
baseline=1cm, trim axis left, trim axis right
]

\begin{axis}[%
width=0.951\figurewidth,
height=\figureheight,
at={(0\figurewidth,0\figureheight)},
scale only axis,
unbounded coords=jump,
separate axis lines,
every outer x axis line/.append style={black},
every x tick label/.append style={font=\color{black}},
every x tick/.append style={black},
xmin=0.5,
xmax=6.5,
xtick={1,2,3,4,5,6},
xticklabels={{dead},{stagnant},{mechanical},{artificial},{inert},{apathetic}},
every outer y axis line/.append style={black},
every y tick label/.append style={font=\color{black}},
every y tick/.append style={black},
ymin=0.85,
ymax=5.15,
ytick={1,2,3,4,5},
yticklabels={\empty},
axis background/.style={fill=white},
ymajorgrids,
xticklabel style={rotate=30, anchor=east, font=\scriptsize, yshift=-0.1cm, xshift=0.05cm}, yticklabel style={font=\scriptsize}
]
\addplot [color=black, dashed, forget plot]
  table[row sep=crcr]{%
1	4.5\\
1	5\\
};
\addplot [color=black, dashed, forget plot]
  table[row sep=crcr]{%
2	4\\
2	5\\
};
\addplot [color=black, dashed, forget plot]
  table[row sep=crcr]{%
3	3\\
3	4\\
};
\addplot [color=black, dashed, forget plot]
  table[row sep=crcr]{%
4	3\\
4	4\\
};
\addplot [color=black, dashed, forget plot]
  table[row sep=crcr]{%
5	4.5\\
5	5\\
};
\addplot [color=black, dashed, forget plot]
  table[row sep=crcr]{%
6	5\\
6	5\\
};
\addplot [color=black, dashed, forget plot]
  table[row sep=crcr]{%
1	2\\
1	3\\
};
\addplot [color=black, dashed, forget plot]
  table[row sep=crcr]{%
2	2\\
2	3\\
};
\addplot [color=black, dashed, forget plot]
  table[row sep=crcr]{%
3	1\\
3	2\\
};
\addplot [color=black, dashed, forget plot]
  table[row sep=crcr]{%
4	1\\
4	2\\
};
\addplot [color=black, dashed, forget plot]
  table[row sep=crcr]{%
5	3\\
5	3\\
};
\addplot [color=black, dashed, forget plot]
  table[row sep=crcr]{%
6	4\\
6	4\\
};
\addplot [color=black, forget plot]
  table[row sep=crcr]{%
0.875	5\\
1.125	5\\
};
\addplot [color=black, forget plot]
  table[row sep=crcr]{%
1.875	5\\
2.125	5\\
};
\addplot [color=black, forget plot]
  table[row sep=crcr]{%
2.875	4\\
3.125	4\\
};
\addplot [color=black, forget plot]
  table[row sep=crcr]{%
3.875	4\\
4.125	4\\
};
\addplot [color=black, forget plot]
  table[row sep=crcr]{%
4.875	5\\
5.125	5\\
};
\addplot [color=black, forget plot]
  table[row sep=crcr]{%
5.875	5\\
6.125	5\\
};
\addplot [color=black, forget plot]
  table[row sep=crcr]{%
0.875	2\\
1.125	2\\
};
\addplot [color=black, forget plot]
  table[row sep=crcr]{%
1.875	2\\
2.125	2\\
};
\addplot [color=black, forget plot]
  table[row sep=crcr]{%
2.875	1\\
3.125	1\\
};
\addplot [color=black, forget plot]
  table[row sep=crcr]{%
3.875	1\\
4.125	1\\
};
\addplot [color=black, forget plot]
  table[row sep=crcr]{%
4.875	3\\
5.125	3\\
};
\addplot [color=black, forget plot]
  table[row sep=crcr]{%
5.875	4\\
6.125	4\\
};
\addplot [color=blue, forget plot]
  table[row sep=crcr]{%
0.75	3\\
0.75	4.5\\
1.25	4.5\\
1.25	3\\
0.75	3\\
};
\addplot [color=blue, forget plot]
  table[row sep=crcr]{%
1.75	3\\
1.75	4\\
2.25	4\\
2.25	3\\
1.75	3\\
};
\addplot [color=blue, forget plot]
  table[row sep=crcr]{%
2.75	2\\
2.75	3\\
3.25	3\\
3.25	2\\
2.75	2\\
};
\addplot [color=blue, forget plot]
  table[row sep=crcr]{%
3.75	2\\
3.75	3\\
4.25	3\\
4.25	2\\
3.75	2\\
};
\addplot [color=blue, forget plot]
  table[row sep=crcr]{%
4.75	3\\
4.75	4.5\\
5.25	4.5\\
5.25	3\\
4.75	3\\
};
\addplot [color=blue, forget plot]
  table[row sep=crcr]{%
5.75	4\\
5.75	5\\
6.25	5\\
6.25	4\\
5.75	4\\
};
\addplot [color=red, line width=1.5pt, forget plot]
  table[row sep=crcr]{%
0.75	4\\
1.25	4\\
};
\addplot [color=red, line width=1.5pt, forget plot]
  table[row sep=crcr]{%
1.75	3.5\\
2.25	3.5\\
};
\addplot [color=red, line width=1.5pt, forget plot]
  table[row sep=crcr]{%
2.75	2\\
3.25	2\\
};
\addplot [color=red, line width=1.5pt, forget plot]
  table[row sep=crcr]{%
3.75	2.5\\
4.25	2.5\\
};
\addplot [color=red, line width=1.5pt, forget plot]
  table[row sep=crcr]{%
4.75	4\\
5.25	4\\
};
\addplot [color=red, line width=1.5pt, forget plot]
  table[row sep=crcr]{%
5.75	4\\
6.25	4\\
};
\addplot [color=black, only marks, mark=+, mark options={solid, draw=blue}, forget plot]
  table[row sep=crcr]{%
nan	nan\\
};
\addplot [color=black, only marks, mark=+, mark options={solid, draw=blue}, forget plot]
  table[row sep=crcr]{%
nan	nan\\
};
\addplot [color=black, only marks, mark=+, mark options={solid, draw=blue}, forget plot]
  table[row sep=crcr]{%
nan	nan\\
};
\addplot [color=black, only marks, mark=+, mark options={solid, draw=blue}, forget plot]
  table[row sep=crcr]{%
nan	nan\\
};
\addplot [color=black, only marks, mark=+, mark options={solid, draw=blue}, forget plot]
  table[row sep=crcr]{%
nan	nan\\
};
\addplot [color=black, only marks, mark=+, mark options={solid, draw=blue}, forget plot]
  table[row sep=crcr]{%
nan	nan\\
};
\end{axis}

\begin{axis}[%
width=0.951\figurewidth,
height=\figureheight,
at={(0\figurewidth,0\figureheight)},
scale only axis,
every outer x axis line/.append style={black},
every x tick label/.append style={font=\color{black}},
every x tick/.append style={black},
xmin=0.5,
xmax=6.5,
xtick={1,2,3,4,5,6},
xticklabels={{alive},{lively},{organic},{lifelike},{interactive},{responsive}},
every outer y axis line/.append style={none},
every y tick label/.append style={font=\color{none}},
every y tick/.append style={none},
ymin=0.85,
ymax=5.15,
ytick={1,1.5,2,2.5,3,3.5,4,4.5,5},
yticklabels={\empty},
axis x line*=top,
axis y line*=left,
xticklabel style={rotate=30, anchor=west, font=\scriptsize, yshift=0.15cm, xshift=-0.1cm}, yticklabel style={font=\scriptsize}
]
\end{axis}
\end{tikzpicture}%
		%	\caption{Animacy \\ ($\alpha=0.66$)}
	\end{subfigure}
	\hfill
	\begin{subfigure}[t]{0.833\sfwidth}
		\setlength\figurewidth{\textwidth}
\begin{tikzpicture}[%
baseline=1cm, trim axis left, trim axis right
]

\begin{axis}[%
width=0.951\figurewidth,
height=\figureheight,
at={(0\figurewidth,0\figureheight)},
scale only axis,
separate axis lines,
every outer x axis line/.append style={black},
every x tick label/.append style={font=\color{black}},
every x tick/.append style={black},
xmin=0.5,
xmax=5.5,
xtick={1,2,3,4,5},
xticklabels={{incompetent},{ignorant},{irresponsible},{unintelligent},{foolish}},
every outer y axis line/.append style={black},
every y tick label/.append style={font=\color{black}},
every y tick/.append style={black},
ymin=0.85,
ymax=5.15,
ytick={1,2,3,4,5},
yticklabels={\empty},
axis background/.style={fill=white},
ymajorgrids,
xticklabel style={rotate=30, anchor=east, font=\scriptsize, yshift=-0.1cm, xshift=0.05cm}, yticklabel style={font=\scriptsize}
]
\addplot [color=black, dashed, forget plot]
  table[row sep=crcr]{%
1	4\\
1	5\\
};
\addplot [color=black, dashed, forget plot]
  table[row sep=crcr]{%
2	4\\
2	5\\
};
\addplot [color=black, dashed, forget plot]
  table[row sep=crcr]{%
3	4\\
3	5\\
};
\addplot [color=black, dashed, forget plot]
  table[row sep=crcr]{%
4	4\\
4	4\\
};
\addplot [color=black, dashed, forget plot]
  table[row sep=crcr]{%
5	5\\
5	5\\
};
\addplot [color=black, dashed, forget plot]
  table[row sep=crcr]{%
1	3\\
1	3\\
};
\addplot [color=black, dashed, forget plot]
  table[row sep=crcr]{%
2	2\\
2	3\\
};
\addplot [color=black, dashed, forget plot]
  table[row sep=crcr]{%
3	2\\
3	3\\
};
\addplot [color=black, dashed, forget plot]
  table[row sep=crcr]{%
4	3\\
4	3\\
};
\addplot [color=black, dashed, forget plot]
  table[row sep=crcr]{%
5	4\\
5	4\\
};
\addplot [color=black, forget plot]
  table[row sep=crcr]{%
0.875	5\\
1.125	5\\
};
\addplot [color=black, forget plot]
  table[row sep=crcr]{%
1.875	5\\
2.125	5\\
};
\addplot [color=black, forget plot]
  table[row sep=crcr]{%
2.875	5\\
3.125	5\\
};
\addplot [color=black, forget plot]
  table[row sep=crcr]{%
3.875	4\\
4.125	4\\
};
\addplot [color=black, forget plot]
  table[row sep=crcr]{%
4.875	5\\
5.125	5\\
};
\addplot [color=black, forget plot]
  table[row sep=crcr]{%
0.875	3\\
1.125	3\\
};
\addplot [color=black, forget plot]
  table[row sep=crcr]{%
1.875	2\\
2.125	2\\
};
\addplot [color=black, forget plot]
  table[row sep=crcr]{%
2.875	2\\
3.125	2\\
};
\addplot [color=black, forget plot]
  table[row sep=crcr]{%
3.875	3\\
4.125	3\\
};
\addplot [color=black, forget plot]
  table[row sep=crcr]{%
4.875	4\\
5.125	4\\
};
\addplot [color=blue, forget plot]
  table[row sep=crcr]{%
0.75	3\\
0.75	4\\
1.25	4\\
1.25	3\\
0.75	3\\
};
\addplot [color=blue, forget plot]
  table[row sep=crcr]{%
1.75	3\\
1.75	4\\
2.25	4\\
2.25	3\\
1.75	3\\
};
\addplot [color=blue, forget plot]
  table[row sep=crcr]{%
2.75	3\\
2.75	4\\
3.25	4\\
3.25	3\\
2.75	3\\
};
\addplot [color=blue, forget plot]
  table[row sep=crcr]{%
3.75	3\\
3.75	4\\
4.25	4\\
4.25	3\\
3.75	3\\
};
\addplot [color=blue, forget plot]
  table[row sep=crcr]{%
4.75	4\\
4.75	5\\
5.25	5\\
5.25	4\\
4.75	4\\
};
\addplot [color=red, line width=1.5pt, forget plot]
  table[row sep=crcr]{%
0.75	3.5\\
1.25	3.5\\
};
\addplot [color=red, line width=1.5pt, forget plot]
  table[row sep=crcr]{%
1.75	4\\
2.25	4\\
};
\addplot [color=red, line width=1.5pt, forget plot]
  table[row sep=crcr]{%
2.75	3.5\\
3.25	3.5\\
};
\addplot [color=red, line width=1.5pt, forget plot]
  table[row sep=crcr]{%
3.75	4\\
4.25	4\\
};
\addplot [color=red, line width=1.5pt, forget plot]
  table[row sep=crcr]{%
4.75	4.5\\
5.25	4.5\\
};
\addplot [color=black, only marks, mark=+, mark options={solid, draw=blue}, forget plot]
  table[row sep=crcr]{%
1	1\\
};
\addplot [color=black, only marks, mark=+, mark options={solid, draw=blue}, forget plot]
  table[row sep=crcr]{%
2	1\\
};
\addplot [color=black, only marks, mark=+, mark options={solid, draw=blue}, forget plot]
  table[row sep=crcr]{%
3	1\\
};
\addplot [color=black, only marks, mark=+, mark options={solid, draw=blue}, forget plot]
  table[row sep=crcr]{%
4	1\\
};
\addplot [color=black, only marks, mark=+, mark options={solid, draw=blue}, forget plot]
  table[row sep=crcr]{%
5	1\\
};
\end{axis}

\begin{axis}[%
width=0.951\figurewidth,
height=\figureheight,
at={(0\figurewidth,0\figureheight)},
scale only axis,
every outer x axis line/.append style={black},
every x tick label/.append style={font=\color{black}},
every x tick/.append style={black},
xmin=0.5,
xmax=5.5,
xtick={1,2,3,4,5},
xticklabels={{competent},{knowledgeable},{responsible},{intelligent},{sensible}},
every outer y axis line/.append style={none},
every y tick label/.append style={font=\color{none}},
every y tick/.append style={none},
ymin=0.85,
ymax=5.15,
ytick={1,1.5,2,2.5,3,3.5,4,4.5,5},
yticklabels={\empty},
axis x line*=top,
axis y line*=left,
xticklabel style={rotate=30, anchor=west, font=\scriptsize, yshift=0.15cm, xshift=-0.1cm}, yticklabel style={font=\scriptsize}
]
\end{axis}
\end{tikzpicture}%
		%	\caption{Perceived Intelligence \\ ($\alpha=0.90$)}
	\end{subfigure}
	\hfill
	\begin{subfigure}[t]{0.5\sfwidth}
		\setlength\figurewidth{\textwidth}
\begin{tikzpicture}[%
baseline=1cm, trim axis left, trim axis right
]

\begin{axis}[%
width=0.951\figurewidth,
height=\figureheight,
at={(0\figurewidth,0\figureheight)},
scale only axis,
unbounded coords=jump,
separate axis lines,
every outer x axis line/.append style={black},
every x tick label/.append style={font=\color{black}},
every x tick/.append style={black},
xmin=0.5,
xmax=3.5,
xtick={1,2,3},
xticklabels={{anxious},{quiescent},{agitated}},
every outer y axis line/.append style={black},
every y tick label/.append style={font=\color{black}},
every y tick/.append style={black},
ymin=0.85,
ymax=5.15,
ytick={1,2,3,4,5},
yticklabels={\empty},
axis background/.style={fill=white},
ymajorgrids,
xticklabel style={rotate=30, anchor=east, font=\scriptsize, yshift=-0.1cm, xshift=0.05cm}, yticklabel style={font=\scriptsize}
]
\addplot [color=black, dashed, forget plot]
  table[row sep=crcr]{%
1	5\\
1	5\\
};
\addplot [color=black, dashed, forget plot]
  table[row sep=crcr]{%
2	4\\
2	5\\
};
\addplot [color=black, dashed, forget plot]
  table[row sep=crcr]{%
3	5\\
3	5\\
};
\addplot [color=black, dashed, forget plot]
  table[row sep=crcr]{%
1	4\\
1	4.5\\
};
\addplot [color=black, dashed, forget plot]
  table[row sep=crcr]{%
2	1\\
2	2.5\\
};
\addplot [color=black, dashed, forget plot]
  table[row sep=crcr]{%
3	4\\
3	4\\
};
\addplot [color=black, forget plot]
  table[row sep=crcr]{%
0.8875	5\\
1.1125	5\\
};
\addplot [color=black, forget plot]
  table[row sep=crcr]{%
1.8875	5\\
2.1125	5\\
};
\addplot [color=black, forget plot]
  table[row sep=crcr]{%
2.8875	5\\
3.1125	5\\
};
\addplot [color=black, forget plot]
  table[row sep=crcr]{%
0.8875	4\\
1.1125	4\\
};
\addplot [color=black, forget plot]
  table[row sep=crcr]{%
1.8875	1\\
2.1125	1\\
};
\addplot [color=black, forget plot]
  table[row sep=crcr]{%
2.8875	4\\
3.1125	4\\
};
\addplot [color=blue, forget plot]
  table[row sep=crcr]{%
0.775	4.5\\
0.775	5\\
1.225	5\\
1.225	4.5\\
0.775	4.5\\
};
\addplot [color=blue, forget plot]
  table[row sep=crcr]{%
1.775	2.5\\
1.775	4\\
2.225	4\\
2.225	2.5\\
1.775	2.5\\
};
\addplot [color=blue, forget plot]
  table[row sep=crcr]{%
2.775	4\\
2.775	5\\
3.225	5\\
3.225	4\\
2.775	4\\
};
\addplot [color=red, line width=1.5pt, forget plot]
  table[row sep=crcr]{%
0.775	5\\
1.225	5\\
};
\addplot [color=red, line width=1.5pt, forget plot]
  table[row sep=crcr]{%
1.775	4\\
2.225	4\\
};
\addplot [color=red, line width=1.5pt, forget plot]
  table[row sep=crcr]{%
2.775	4\\
3.225	4\\
};
\addplot [color=black, only marks, mark=+, mark options={solid, draw=blue}, forget plot]
  table[row sep=crcr]{%
1	2\\
1	2\\
};
\addplot [color=black, only marks, mark=+, mark options={solid, draw=blue}, forget plot]
  table[row sep=crcr]{%
nan	nan\\
};
\addplot [color=black, only marks, mark=+, mark options={solid, draw=blue}, forget plot]
  table[row sep=crcr]{%
3	2\\
};
\end{axis}

\begin{axis}[%
width=0.951\figurewidth,
height=\figureheight,
at={(0\figurewidth,0\figureheight)},
scale only axis,
every outer x axis line/.append style={black},
every x tick label/.append style={font=\color{black}},
every x tick/.append style={black},
xmin=0.5,
xmax=3.5,
xtick={1,2,3},
xticklabels={{relaxed},{surprised},{calm}},
every outer y axis line/.append style={none},
every y tick label/.append style={font=\color{none}},
every y tick/.append style={none},
ymin=0.85,
ymax=5.15,
ytick={1,1.5,2,2.5,3,3.5,4,4.5,5},
yticklabels={\empty},
axis x line*=top,
axis y line*=left,
xticklabel style={rotate=30, anchor=west, font=\scriptsize, yshift=0.15cm, xshift=-0.1cm}, yticklabel style={font=\scriptsize}
]
\end{axis}
\end{tikzpicture}%
		%	\caption{Perceived Safety \\ ($\alpha=-0.20$)}
	\end{subfigure}
	
	\begin{subfigure}[t]{0.78\sfwidth}
		\caption{Anthropomorphism \\ ($\alpha=0.84$)}
	\end{subfigure}
	\hspace{10mm}
	\begin{subfigure}[t]{0.5\sfwidth}
		\caption{Likeability \\ ($\alpha=0.83$)}
	\end{subfigure}
	\hfill
	\begin{subfigure}[t]{0.5\sfwidth}
		\caption{Animacy \\ ($\alpha=0.66$)}
	\end{subfigure}
	\hspace{13mm}
	\begin{subfigure}[t]{0.7\sfwidth}
		\centering
		\caption{Perceived Intelligence \\ ($\alpha=0.90$)}
	\end{subfigure}
	\hspace{0mm}
	\begin{subfigure}[t]{0.55\sfwidth}
		\centering
		\caption{Perceived Safety \\ ($\alpha=-0.20$)}
	\end{subfigure}
	\vspace{-3mm}
	\caption{Results of the user study on sensations and perceptions (\acl{gqs}) of a user during interaction with Sobi.
		The consistency within a category is indicated in each case with Cronbach's $\alpha$. The values for the anthropomorphism, likeability, and perceived intelligence categories are above the 0.70 threshold, indicating internal consistency. A low value in the safety category was also reported in similar research \cite{Tan2018}.}
	\label{fig:godspeed} 
	\vspace{-3mm}
\end{figure*}
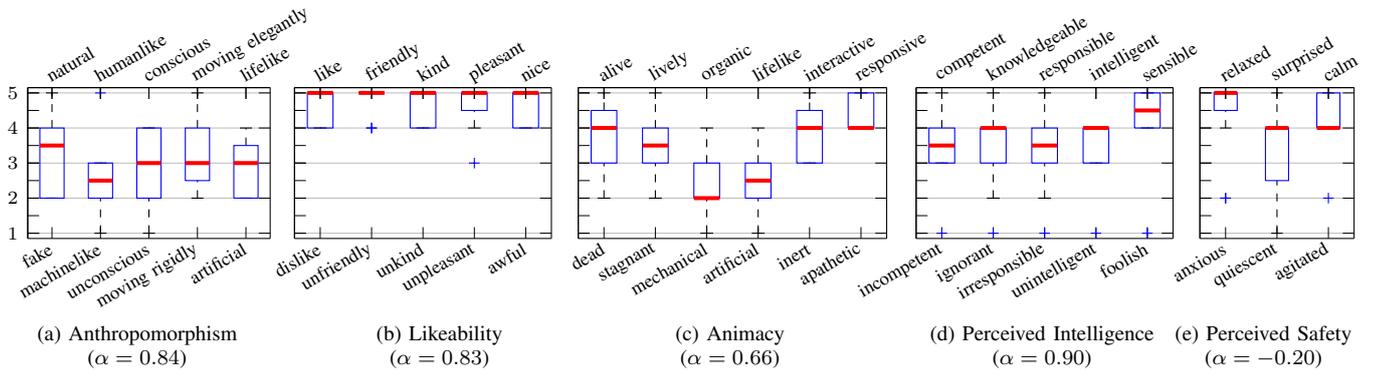
The results of the GQS  are shown in Fig.~\ref{fig:godspeed}. 
For the evaluation of reliability, the respective value of Cronbach's alpha represents a measure of internal consistency within a category. 
The results show that the robot likeability level felt during the interaction is not dependent on distinctive human-like characteristics of the robot. 
Even though human-like characteristics within the anthropomorphism category can only be partially applied to Sobi, the interaction was predominantly perceived as pleasant. 
Sobi's appearance and behavior are positively highlighted with overall high perceived likability by all users. 
%Within the animacy category, internal consistency is low. 
Sobi is perceived as a mechanical robot, but its dynamic behavior clearly distinguishes him from a static machine. 
By greeting a user automatically when he or she approaches as well as providing immediate verbal and visual feedback to user requests, the robot is perceived as lively and responsive.
With high internal consistency within the corresponding category, the perceived intelligence of the robot was rated relatively positively with medium to high values, but continues to show potential for development. 
In many cases, speech input was used to search for locations, which were not always correctly comprehended by the robot or were fundamentally unknown to it. 
However, regardless of any failed attempts, the robot's behavior in responding to user instructions is considered reasonable. 
The outliers at values of 1 could be due to an inadvertent erroneous rating of the category by one person, since this person otherwise positively highlighted the interaction with Sobi as well as his abilities in an additional free text field that had to be answered. 
Other people described the operation of the robot as smooth and intuitive, but also mentioned possible scopes of knowledge that could be learned by Sobi. 
Overall, Sobi was perceived as a competent robot, which was able to answer most of the users' concerns. 
Due to its appearance as well as its active and reactive behavior, the robot was accepted as a pleasant interaction partner.

	    \section{Lessons learned}
	\label{sec:lessons}
	
During the development period and the evaluation runs we had the opportunity to test Sobi, identify errors and obtain feedback.
We will therefore briefly discuss the most important insights and challenges of this process.
The most prominent factor in the first months of development of the robot was the choice of main hardware components (i.e. the mobile base, sensors and computers).
Since the selection of reliable hardware components is the basis for long-term autonomous use, we mainly use established components with maintained ROS support, full integration and standard transmission interfaces (e.g. USB or Ethernet).
The signal lines are kept as short and shielded as possible, since in our experience, especially with data-intensive transmission (e.g. camera data), line losses or interference lead to errors as the operating time increases.
To develop the SLAM functionality of the robot, a functional model with a simple aluminum frame structure was used early in the development process.
This allowed the exact positioning of sensors to be easily varied.
After these initial tests, we started working with the design team early on to conceptualize and develop the eventual design of the robot appearance.
By defining this at an early stage, it was thus possible to avoid redesigns of the mechanical structure and provide a guideline for the placement of the other robot components.

The robot initially contained only the main computer and the computer in the mobile platform to run the software.
However, it has proven useful to outsource individual functions to function-specific embedded computers.
The execution of many hardware drivers can already lead to a considerable system load, which can be counteracted by distributing them over several computing units.
Thus, for example, the camera data is read out and synchronized by the Jetson computer, so that only the compressed RGBD data has to be made available to the ROS system via Ethernet.
An essential prerequisite for this is that the system clock times of all computers are synchronized with each other to avoid inconsistencies in ROS message processing.
A common approach is to use \acf{ntp}, with the main computer acting as the \ac{ntp} server in our case, so time synchronization is possible regardless of an internet connection.

In terms of software development and implementation for long term autonomy, we share the  findings from similar efforts in the literature \cite{Meeussen2011,Hawes2016,Wang2018 }.
Especially the design principles that all programs should tolerate the temporary unavailability of other programs and that program restarts should transition to a clean state are the basis for the design of our software and the monitoring system.
In terms of localization, our system differs from the aforementioned literature in the use of multiple RGBD cameras in conjunction with a 3D laser scanner.
Since the localization programs are terminated during the charging process, it is necessary to quickly regain a valid localization estimate when resuming operation.
For this purpose, we hung up a poster with many images and text directly above the charging station, so that a large number of distinctive features leads to a loop closure detection directly after undocking (see Fig.~\ref{fig:environment}).
It has also proven useful to cover a large field of view with the different perceptual sensors in order to prevent the robot from losing localization if, for example, people are standing around it.
On the other hand, the fields of view of the cameras and the 3D laser scanner overlap, so that the depth information of the scanner can be registered with the RGB images.
This is particularly advantageous for large distances, since the depth data from the cameras is only useful for distances of a few meters.
	
We also had to discard some original plans during the course of the project.
The implementation of actuators requires a significant amount of time and therefore we did not implement a rotatable head and a tiltable tablet as originally intended.
Furthermore, the plan to let the robot move autonomously between indoor and outdoor areas is prevented by door sills.
These are too high (20--30\SI{}{\milli\meter}) for the robot to pass, due to its mass of \SI{150}{\kilogram} with a wheel diameter of \SI{260}{\milli\meter}.
This problem arises from an incorrect assumption regarding accessibility, as the campus was still under construction during the project period.
    \section{Conclusion and Future Works}
    	\label{sec:conclusion}
    	In this work we presented our approach to building a long-term autonomous robot in terms of design, hardware components and software structure.
	%We presented Sobi, a mobile service robot to provide information and guiding for indoor and outdoor environments, as a fully open source platform.
	The robot's monitoring system was presented and tested in a 16-day evaluation with regard to long-term autonomy.
	In a user study with 12 participants, the functionalities of the \ac{hri} were assessed in terms of impression and usefulness.
	Participants described the services offered by the robot as useful and perceived the appearance as very pleasant.
	During the long-term evaluation the robot was in motion for \SI{65.4}{\hour}, traveling a total distance of \SI{66.6}{\kilo\meter}.
	These first results show that the robot is suitable for long-term autonomous tasks, since the main sources of error were due to the developed software.
	In addition to correcting these errors, the robot will be tested in further outdoor deployments in the future to test the capabilites for long-term autonomy under even more dynamic environmental conditions.
	Within the deployment on our campus, further insights are to be gained through a larger number of users and recording of time- and location-dependent long-term usage patterns.
	This information will then be used to create data-driven models of people occurrence and usage patterns, which could then be employed to actively improve the services offered by the robot.

		\section*{Acknowledgment}
	This work was funded by the Faculty of Mechanical Engineering of the Leibniz University Hannover. Special thanks go to the product design team of the Hannover University of Applied Sciences and Arts.
	\bibliography{library}
	\bibliographystyle{IEEEtran}

	\end{document}